\documentclass[a4paper,fleqn]{cas-dc}

\usepackage{amsmath,amsfonts,epsfig}
\usepackage{algorithmic}
\usepackage{array}
\usepackage[caption=false,font=normalsize,labelfont=sf,textfont=sf]{subfig}
\usepackage{textcomp}
\usepackage{stfloats}
\usepackage{url}
\usepackage{verbatim}
\usepackage{enumitem}
\usepackage{amssymb}
\usepackage{capt-of} % provides \captionof without overriding CAS captions
\usepackage{booktabs}
\usepackage[export]{adjustbox}
\usepackage{xspace}
\usepackage{graphicx}
\usepackage{comment}
\usepackage{colortbl}
\usepackage{tcolorbox}
\usepackage{etoolbox}
\usepackage{tikz}
\usetikzlibrary{patterns}
\usepackage{pgf}
\usepackage{makecell}
\usepackage[table,dvipsnames]{xcolor}
\definecolor{cvprblue}{rgb}{0.21,0.49,0.74}
\hypersetup{
  colorlinks=true,
  citecolor=cvprblue,
  linkcolor=cvprblue,
  urlcolor=cvprblue
}
\usepackage{multirow}
\usepackage{setspace}
\usepackage{orcidlink}
\usepackage{balance}
\usepackage{pgfplots}
\pgfplotsset{compat=1.18}
\usepackage{diagbox}
\usepackage{pifont}
\usepackage{longtable}
\usepackage{lineno}
\usepackage[numbers]{natbib}
\usepackage{float}

\newcommand{\Th}[1]{\textsc{#1}}

\def\l1{\ensuremath{\ell_1}\xspace}
\def\l2{\ensuremath{\ell_2}\xspace}

\newcommand{\cI}{\mathcal{I}}

\newcommand{\cT}{\mathcal{T}}

\makeatletter
\DeclareRobustCommand\onedot{\futurelet\@let@token\@onedot}
\def\@onedot{\ifx\@let@token.\else.\null\fi\xspace}
\def\eg{\emph{e.g}\onedot}

\makeatother

\ExplSyntaxOn

\NewDocumentEnvironment{casfigurehere}{ O{} }
 {
  \par\addvspace{6pt}
  \group_begin:
  \__reset_fig:
  \bool_gset_false:N \g_fig_full_bool
  \dim_set:Nn \l_fig_width_dim { \linewidth }
  \keys_set:nn { cas / fig } { #1 }
  \l_fig_align_tl
  \sffamily\small
 }
 {
  \par
  \group_end:
  \addvspace{6pt}
 }

\NewDocumentCommand{\casfigcaption}{ m +m }
 {
  \par
  \refstepcounter{figure}
  \__make_fig_caption:nn { \figurename\space\thefigure } { #2 }
  \label{#1}
  \par
 }

\NewDocumentEnvironment{castablehere}{ O{} }
 {
  \par\addvspace{6pt}
  \group_begin:
  \__reset_tbl:
  \bool_gset_false:N \g_tbl_full_bool
  \dim_set:Nn \l_tbl_width_dim { \linewidth }
  \keys_set:nn { cas / tbl } { #1 }
  \l_tbl_align_tl
  \sffamily\small
 }
 {
  \par
  \group_end:
  \addvspace{6pt}
 }

\NewDocumentCommand{\castablecaption}{ m +m }
 {
  \par
  \refstepcounter{table}
  \__make_tbl_caption:nn { \tablename\space\thetable } { #2 }
  \label{#1}
  \par
 }

\ExplSyntaxOff
\definecolor{appleblue}{RGB}{80,122,255}  % Softer, muted blue
\definecolor{applepink}{RGB}{217,127,174}  % Muted, soft pink
\definecolor{appleteal}{RGB}{85,163,152}  % Soft, calming teal
\definecolor{applepurple}{RGB}{138,107,221}  % Muted, soft purple
\definecolor{applemustard}{RGB}{234,179,77}  % Warm, soft mustard yellow
\definecolor{appleorange}{RGB}{255,127,80}  % Vibrant, distinctive orange
\definecolor{applegreen}{RGB}{52,199,89}
\definecolor{appleyellow}{RGB}{255,204,0}
\definecolor{applered}{RGB}{255,59,48}  % Warm, vibrant apple red

\newcommand{\cmark}{\textcolor{green!60!black}{\ding{51}}} % ✓
\newcommand{\xmark}{\textcolor{red!75!black}{\ding{55}}}  % ✗

\definecolor{OrangeFrame}{HTML}{D79B00}

\definecolor{textquerycolor}{RGB}{16,115,158}
\definecolor{retrievedimagecolor}{RGB}{150,115,166}

\definecolor{LightSteelBlue1}{RGB}{202, 225, 255}
\definecolor{high}{HTML}{76f013}  % the color for the highest number in your data set
\definecolor{low}{HTML}{ec462e}  % the color for the lowest number in your data set
\newcommand*{\opacity}{60}% here you can change the opacity of the background color!
\newcommand*{\minvalcolor}{14.5}% define the minimum value on your data set
\newcommand*{\maxvalcolor}{55.3}% define the maximum value in your data set!
\newcommand{\gradientcolor}[1]{
    \pgfmathparse{(#1-\minvalcolor)/(\maxvalcolor-\minvalcolor)}
    \let\normalizedval\pgfmathresult

    \pgfmathparse{100*(\normalizedval)^(2.0)} % Exponential scaling
    \xdef\tempa{\pgfmathresult}

    \pgfmathparse{min(100,max(0,\tempa))}
    \xdef\tempa{\pgfmathresult}

    \cellcolor{high!\tempa!low!\opacity} #1
}
\newcommand*{\minvalcontext}{6.6}% define the minimum value on your data set
\newcommand*{\maxvalcontext}{31.5}% define the maximum value in your data set!
\newcommand{\gradientcontext}[1]{
    \pgfmathparse{(#1-\minvalcontext)/(\maxvalcontext-\minvalcontext)}
    \let\normalizedval\pgfmathresult

    \pgfmathparse{100*(\normalizedval)^(2.0)} % Exponential scaling
    \xdef\tempa{\pgfmathresult}

    \pgfmathparse{min(100,max(0,\tempa))}
    \xdef\tempa{\pgfmathresult}

    \cellcolor{high!\tempa!low!\opacity} #1
}
\newcommand*{\minvaldensity}{15.1}% define the minimum value on your data set
\newcommand*{\maxvaldensity}{42}% define the maximum value in your data set!
\newcommand{\gradientdensity}[1]{
    \pgfmathparse{(#1-\minvaldensity)/(\maxvaldensity-\minvaldensity)}
    \let\normalizedval\pgfmathresult

    \pgfmathparse{100*(\normalizedval)^(3.0)} % Exponential scaling
    \xdef\tempa{\pgfmathresult}

    \pgfmathparse{min(100,max(0,\tempa))}
    \xdef\tempa{\pgfmathresult}

    \cellcolor{high!\tempa!low!\opacity} #1
}
\newcommand*{\minvalexistence}{7}% define the minimum value on your data set
\newcommand*{\maxvalexistence}{15}% define the maximum value in your data set!
\newcommand{\gradientexistence}[1]{
    \pgfmathparse{(#1-\minvalexistence)/(\maxvalexistence-\minvalexistence)}
    \let\normalizedval\pgfmathresult

    \pgfmathparse{100*(\normalizedval)^(4)} % Exponential scaling
    \xdef\tempa{\pgfmathresult}

    \pgfmathparse{min(100,max(0,\tempa))}
    \xdef\tempa{\pgfmathresult}

    \cellcolor{high!\tempa!low!\opacity} #1
}
\newcommand*{\minvalquantity}{7.0}% define the minimum value on your data set
\newcommand*{\maxvalquantity}{20.9}% define the maximum value in your data set!
\newcommand{\gradientquantity}[1]{
    \pgfmathparse{(#1-\minvalquantity)/(\maxvalquantity-\minvalquantity)}
    \let\normalizedval\pgfmathresult

    \pgfmathparse{100*(\normalizedval)^(4.0)} % Exponential scaling
    \xdef\tempa{\pgfmathresult}

    \pgfmathparse{min(100,max(0,\tempa))}
    \xdef\tempa{\pgfmathresult}

    \cellcolor{high!\tempa!low!\opacity} #1
}
\newcommand*{\minvalshape}{15.2}% define the minimum value on your data set
\newcommand*{\maxvalshape}{32}% define the maximum value in your data set!
\newcommand{\gradientshape}[1]{
    \pgfmathparse{(#1-\minvalshape)/(\maxvalshape-\minvalshape)}
    \let\normalizedval\pgfmathresult

    \pgfmathparse{100*(\normalizedval)^(8.0)} % Exponential scaling
    \xdef\tempa{\pgfmathresult}

    \pgfmathparse{min(100,max(0,\tempa))}
    \xdef\tempa{\pgfmathresult}

    \cellcolor{high!\tempa!low!\opacity} #1
}
\newcommand*{\minvalaverage}{11.3}% define the minimum value on your data set
\newcommand*{\maxvalaverage}{30.2}% define the maximum value in your data set!
\newcommand{\gradientaverage}[1]{
    \pgfmathparse{(#1-\minvalaverage)/(\maxvalaverage-\minvalaverage)}
    \let\normalizedval\pgfmathresult

    \pgfmathparse{100*(\normalizedval)^(5.0)} % Exponential scaling
    \xdef\tempa{\pgfmathresult}

    \pgfmathparse{min(100,max(0,\tempa))}
    \xdef\tempa{\pgfmathresult}

    \cellcolor{high!\tempa!low!\opacity} #1
}
\newcommand*{\minvalfreedommk}{26.71}% define the minimum value on your data set
\newcommand*{\maxvalfreedommk}{35.68}% define the maximum value in your data set!
\newcommand{\gradientfreedommk}[1]{
    \pgfmathparse{(#1-\minvalfreedommk)/(\minvalfreedommk-\maxvalfreedommk)}
    \let\normalizedval\pgfmathresult

    \pgfmathparse{100*(\normalizedval)^(2.0)} % Exponential scaling
    \xdef\tempa{\pgfmathresult}

    \pgfmathparse{min(100,max(0,\tempa))}
    \xdef\tempa{\pgfmathresult}

    \cellcolor{high!\tempa!low!\opacity} #1
}
\newcommand*{\minvalfreedomkn}{34.89}% define the minimum value on your data set
\newcommand*{\maxvalfreedomkn}{40.75}% define the maximum value in your data set!
\newcommand{\gradientfreedomkn}[1]{
    \pgfmathparse{(#1-\minvalfreedomkn)/(\minvalfreedomkn-\maxvalfreedomkn)}
    \let\normalizedval\pgfmathresult

    \pgfmathparse{100*(\normalizedval)^(2.0)} % Exponential scaling
    \xdef\tempa{\pgfmathresult}

    \pgfmathparse{min(100,max(0,\tempa))}
    \xdef\tempa{\pgfmathresult}

    \cellcolor{high!\tempa!low!\opacity} #1
}

\newcommand{\best}[1]{\textbf{#1}}
\newcommand{\second}[1]{\underline{#1}}

\newcommand{\qtext}[1]{%
  \begingroup
  \setlength{\fboxsep}{1.4pt}%
  \setlength{\fboxrule}{0.9pt}%
  \fcolorbox{textquerycolor}{white}{\tiny\texttt{\textcolor{textquerycolor}{#1}}}%
  \endgroup
}

\newcommand{\qcell}[2]{%
  \begin{tabular}[t]{@{}c@{}}%
    \includegraphics[width=\sz,height=\sz]{#1}\\[-1pt]
    {\tiny\textbf{+}}\hspace{1pt}\qtext{#2}%
  \end{tabular}%
}

\begin{document}
%\let\WriteBookmarks\relax
%\def\floatpagepagefraction{1}
%\def\textpagefraction{.001}
%\linenumbers

\shorttitle{Benchmarking Composed Image Retrieval for Applied Earth Observation}
\shortauthors{Psomas et~al.}

\title[mode=title]{Benchmarking Composed Image Retrieval \\ for Applied Earth Observation}

\author[ctu]{Bill~Psomas}[orcid=0000-0001-5381-0312]
\author[ntua]{Dionysis~Christopoulos}
\author[ntua]{Thanasis~Petropoulos}
\author[ctu]{Nikos~Efthymiadis}
\author[demokritos]{Ioannis~Kakogeorgiou}
\author[ctu]{Ond\v{r}ej~Chum}
\author[nkuoa]{Yannis~Avrithis}
\author[ctu]{Giorgos~Tolias}
\author[ntua]{Konstantinos~Karantzalos}

\affiliation[ctu]{organization={Visual Recognition Group, Department of Cybernetics, Czech Technical University in Prague}, country={Czechia}}
\affiliation[ntua]{organization={Remote Sensing Laboratory, School of Rural, Surveying and Geoinformatics Engineering, National Technical University of Athens}, country={Greece}}
\affiliation[demokritos]{organization={Institute of Informatics \& Telecommunications, National Centre for Scientific Research ``Demokritos''}, country={Greece}}
\affiliation[nkuoa]{organization={Department of Informatics and Telecommunications, National and Kapodistrian University of Athens}, country={Greece}}

\begin{abstract}
Remote sensing composed image retrieval (RSCIR) enables search in large satellite image archives using composed queries that combine a reference image with a textual modifier. Although RSCIR offers a flexible interface for expressing targeted retrieval intent, the transferability of modern composition methods to Earth observation (EO) imagery and their relevance to operational EO workflows remain underexplored. We address this gap through a unified benchmark and an application-oriented study. First, we systematically adapt and evaluate representative composed image retrieval methods with six vision-language backbones on PatternCom under a standardized protocol, analyzing their behavior across backbones, composition strategies, and query types. Second, we introduce xView2-CIR, a change-centric dataset for disaster and damage monitoring, where retrieval is conditioned on scene identity and a target post-event state. Our results show that training-free composition methods provide strong and scalable baselines for EO retrieval, while change-centric retrieval presents different challenges from attribute-based retrieval, particularly due to the need to preserve scene identity. Overall, this study establishes a practical benchmark for RSCIR and positions composed retrieval as a complementary tool for remote sensing image retrieval, archive exploration, and change analysis.
\end{abstract}


\maketitle

\section{Introduction}
\label{sec:intro}

The explosive growth of earth observation (EO) data has enabled large-scale monitoring of the planet, but has also made it increasingly difficult for users to navigate massive satellite image archives and retrieve content matching specific information needs. 
Remote sensing image retrieval (RSIR)~\cite{agouris} addresses this challenge by searching EO archives from a query, most commonly an image. 
Prior RSIR research spans unisource and cross-source~\cite{zhou2023remote} settings, as well as single-label~\cite{hou2020exploiting, sumbul2022plasticity, cheng2021novel} and multi-label~\cite{kang2020graph, imbriaco2021toward, shao2020multilabel} formulations. 
Despite this progress, a fundamental limitation persists: most RSIR systems are driven by a \emph{single modality} query. 
A user must typically choose between an image query or a text query, which \emph{restricts expressivity} when the retrieval intent includes \emph{targeted modifications} beyond overall similarity.

\begin{figure*}[th]
  \centering
  % adjust width as you like: \textwidth (full) or 0.98\textwidth etc.
  \includegraphics[width=\textwidth]{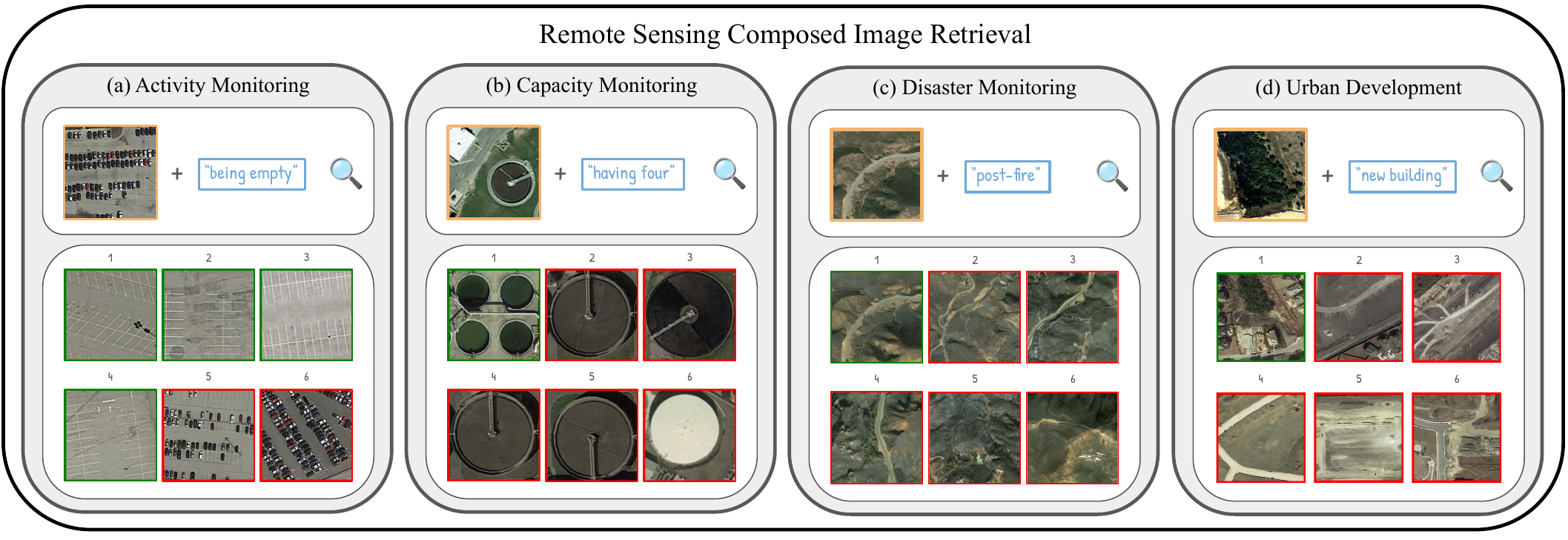}
  \vspace{-16pt}
  \caption{%
    \textbf{Composed Image Retrieval for applied Earth Observation.}
    We illustrate how composed queries (\textcolor{orange}{query image} + \textcolor{textquerycolor}{query text})
    enable controllable retrieval by specifying a targeted change.
    (a) Activity monitoring: a parking-lot query image composed with \texttt{being empty} retrieves visually
    \emph{similar} parking areas with low vehicle occupancy.
    (b) Capacity monitoring: A facility query image composed with \texttt{having four} retrieves visually \emph{similar} sites exhibiting a different quantity attribute.
    (c) Disaster monitoring: a pre-event query image composed with \texttt{post-fire} retrieves post-event imagery of the \emph{same} scene consistent with wildfire impact.
    (d) Urban development: a pre-construction query image composed with \texttt{new buildings} retrieves the \emph{same} scene exhibiting recent construction.
    For each query, retrieved candidates (ranked left-to-right, top-to-bottom) are produced by FreeDom~\cite{freedom} with OpenAI CLIP~\cite{clip}; \textcolor{ForestGreen}{green} borders indicate correct retrievals under the respective evaluation criterion (a,b: same class + target attribute value, c,d: same scene + target change), while \textcolor{red}{red} borders indicate mismatches.
    Images in (a,b) are from PatternNet~\cite{zhou2018patternnet}, (c) from xView2~\cite{xview}, and (d) from LEVIR-CC~\cite{levir}.%
  }
  \label{fig:teaser}
\end{figure*}

Many operational remote sensing workflows are inherently \emph{compositional}. 
For example, an analyst may wish to retrieve locations visually \emph{similar} to a reference region but with a different attribute, such as a parking lot with lower occupancy or a facility with a larger number of storage tanks, as shown in~\autoref{fig:teaser}(a,b). 
In change-centric applications, the goal may instead be to retrieve imagery of the \emph{same} location under a different state, such as after a wildfire or recent construction, as shown in~\autoref{fig:teaser}(c,d). 
Such intent is difficult to express with unimodal queries alone. 
Remote sensing composed image retrieval (RSCIR)~\cite{psomas2024composedimageretrievalremote} addresses this limitation by combining a reference image with a natural-language modifier, aiming to retrieve images that preserve the relevant visual context while satisfying the textual specification~\cite{pic2word,freedom,basic}. 

Despite its promise, RSCIR remains insufficiently studied from an applied EO perspective. Existing work~\cite{psomas2024composedimageretrievalremote} introduced the task, but it remains unclear how modern composed image retrieval methods behave under a unified evaluation protocol, how well they transfer across general-purpose and remote-sensing vision--language backbones, and whether composed retrieval remains effective when relevance depends on scene identity and a target post-event state rather than class-level attribute matching. These questions are important for assessing whether RSCIR can support practical EO workflows such as archive exploration, infrastructure monitoring, disaster response, and urban change analysis.

This work advances RSCIR through a unified benchmark and an application-oriented study. We first benchmark representative composition methods on PatternCom~\cite{psomas2024composedimageretrievalremote}, an attribute-modification testbed derived from PatternNet~\cite{zhou2018patternnet}. We evaluate six vision--language backbones, including CLIP~\cite{clip}, SigLIP~\cite{siglip}, RemoteCLIP~\cite{liu2023remoteclip}, and SkyCLIP~\cite{skyscript2023}, and adapt representative composition methods such as Pic2Word~\cite{pic2word}, SEARLE~\cite{searle}, FreeDom~\cite{freedom}, MagicLens~\cite{zhang2024magiclensselfsupervisedimageretrieval}, and BASIC~\cite{basic} to EO imagery. Beyond aggregate retrieval metrics, we analyze method behavior across backbones, query types, and composition strategies to provide practical guidance for selecting RSCIR pipelines.

To connect benchmarking with operational EO needs, we further introduce xView2-CIR, a new dataset derived from xView2~\cite{xview} for change-centric composed retrieval in disaster and damage monitoring. In xView2-CIR, a query consists of a pre-event reference image and a textual modifier such as \texttt{post-fire}, and the goal is to retrieve the corresponding post-event image of the same location. This setting evaluates whether composed retrieval can support location-aware and semantically steerable search, complementing established change-centric workflows such as change detection, damage assessment, and rapid mapping.

%To bridge benchmarking with real EO use cases, we introduce a new RSCIR dataset derived from xView2~\cite{xview} that targets \emph{disaster and damage monitoring} through composed retrieval. 
%In this setting, the modifier naturally encodes a change-centric intent (\eg, \texttt{post-fire}), enabling retrieval of post-event imagery conditioned on a pre-event reference.
%This dataset supports studying how composed retrieval complements established change-centric workflows (change detection, damage assessment, rapid mapping) by enabling search that is both \emph{location-aware} and \emph{semantically steerable}.

In summary, we make the following contributions:
\begin{enumerate}[itemsep=2pt, parsep=0pt, topsep=3pt]
    \item We establish a unified benchmark for RSCIR on PatternCom, with domain-grounded adaptations of representative composition methods and a standardized protocol spanning six vision-language backbones.
    \item We introduce xView2-CIR, a new evaluation dataset for change-centric composed retrieval in disaster and damage monitoring, where relevance depends on both scene identity and target post-event state.
    \item We provide empirical analysis and practical guidance showing when composed retrieval is useful for remote sensing applications, and how its behavior differs between attribute-based and change-centric settings.
\end{enumerate}
\section{Related Work}
\label{sec:related}

\paragraph{Remote Sensing Image Retrieval.} 
With the aim to effectively \emph{search} and \emph{retrieve} information from extensive remote sensing (RS) image archives, remote sensing image retrieval (RSIR) can be categorized into \emph{unisource} and \emph{cross-source}~\cite{zhou2023remote}.
Initially, RSIR methods focus on handcrafted and low-level visual features~\cite{mamatha2010content, ma2014improved, piedra2013fuzzy, li2004integrated, bhagavathy2006modeling, wang2012remote, shao2014remote, wang2013remote, chaudhuri2017multilabel, dai2017novel}. With the advent of deep learning, neural networks are utilized for unisource \emph{single-label} retrieval: (a) as feature extractors~\cite{li2016content, hu2016delving, boualleg2018enhanced, ye2018remote, napoletano2018visual, ge2018exploiting, tang2018unsupervised, imbriaco2019aggregated, sadeghi2019scalable, hou2020exploiting}, (b) trained from scratch~\cite{zhou2017learning, zhang2021triplet, zhuo2021remote, liu2020remote, wang2021learnable, sumbul2022plasticity, wang2016three, chaudhuri2019siamese}, (c) integrating attention modules~\cite{wang2022novel, wang2020attention, xiong2019discriminative, chaudhuri2021attention} and (d) using metric learning~\cite{zhao2021global, cao2020enhancing, cheng2021novel, liu2020eagle, fan2020global, liu2020similarity}. Neural networks are also used for unisource \emph{multi-label}~\cite{chaudhuri2017multilabel, kang2020graph, sumbul2021novel, sumbul2021informative, cheng2021semantic, imbriaco2021toward, shao2020multilabel}, cross-source \emph{cross-sensors}~\cite{li2018learning, ma2021cross, xiong2020discriminative, xiong2020learning}, cross-source \emph{cross-modal}~\cite{chaudhuri2021attention, xu2020mental, sun2021multisensor, sumbul2021bigearthnet, lv2021fusion, yuan2022remote, yuan2022exploring} and cross-source \emph{cross-view} retrieval~\cite{hu2018cvm, zeng2022geo, tian2020cross, lin2015learning, khurshid2019cross, shi2022beyond}. Our work fills a notable gap and enhances user intent expression in RSIR by combining image with text. 

\paragraph{Composed Image Retrieval.} Image-to-image~\cite{rtc19, gar+16, nas+17} and text-to-image~\cite{sarafianos2019adversarial, zhang2020context, devise} retrieval provide ways to explore large image archives. However, the most accurate and flexible way to express the user intent is a query \emph{composed} of both an image and a text. Composed image retrieval (CIR)~\cite{tirg, val, lbf, combiner, cosmo, pic2word} aims to retrieve images not only visually similar to the query image, but also altered to align with the specifics of the query text. Traditionally, CIR methods are supervised by \emph{triplets} of the form \emph{query image, query text, target image}~\cite{tirg, jvsm, yin2020disentangled, lbf, rhg+15, val, cosmo, artemis}. The labor-intensive process of labeling such triplets limits early works to specific applications in fashion~\cite{fashion200k, shoes, fashioniq}, physical states~\cite{mit_states}, object attributes and composition~\cite{tirg, lmb+14, mpc}. 
The emergence of vision-language models (VLMs) such as CLIP~\cite{clip}, ALIGN~\cite{align}, or BLIP~\cite{blip} has recently enabled \emph{zero-shot composed image retrieval} (ZS-CIR): one can compose a query by manipulating embeddings alone, without any task‑specific training. Early ZS‑CIR attempts include Pic2Word~\cite{pic2word} and SEARLE~\cite{searle}, which invert the visual embedding back to text by either pre‑training a small decoder or optimizing a pseudo token at test‑time respectively. More recent work pushes this idea further with memory‑based inversion (FreeDom~\cite{freedom}) or caption‑and‑LLM pipelines (CIReVL~\cite{cirevl}). CompoDiff~\cite{compodiff} casts the composition as a diffusion~\cite{stablediffusion} problem, gradually blending visual and textual semantics. MagicLens~\cite{zhang2024magiclensselfsupervisedimageretrieval} revisits triplet supervision at web scale, where a VLM with extra attention layers is fine-tuned to project the image-text input to a single embedding. Despite this rapid progress, the extent to which ZS-CIR methods transfer to EO imagery and how they should be benchmarked and used in real-world remote sensing workflows remains insufficiently understood. This paper addresses this gap through a unified benchmark, complemented by an application-oriented study.

\paragraph{Vision-Language Models.} Foundational VLMs such as CLIP~\cite{clip}, ALIGN~\cite{align} and BLIP/BLIP‑2~\cite{blip,blip2} are pre-trained on hundreds of millions, or even billions, of web image–text pairs (\eg LAION‑2B~\cite{laion}). Their joint embedding spaces enable strong zero‑shot transfer to a wide range of tasks, including open‑vocabulary classification~\cite{clip}, detection~\cite{gu2021open}, segmentation~\cite{lposs}, and captioning~\cite{tewel2022zerocap}.  OpenCLIP~\cite{openclip} re‑implements CLIP and releases checkpoints trained directly on LAION‑2B. SigLIP~\cite{siglip} replaces the softmax contrastive loss with a sigmoid variant and is trained on the multilingual WebLI corpus. Several works adapt CLIP to the remote sensing domain: RemoteCLIP~\cite{liu2023remoteclip} fine‑tunes OpenAI CLIP on caption‑augmented RS datasets, CLIP LAION‑RS uses a 726k RS subset of LAION‑2B (LAION-RS), and SkyCLIP~\cite{skyscript2023} leverages SkyScript, a million-scale image-text pair dataset, constructed by linking satellite images from Google Earth Engine~\cite{google_earth_engine} with OpenStreetMap~\cite{openstreetmap} annotations. In this work, we benchmark various VLMs under identical composed image retrieval protocols, revealing how generic versus RS-specialised pre‑training affects the performance on the proposed RSCIR task.
\section{Task \& Methodology}
\label{sec:method}

%\begin{figure*}[t]
%\centering
%\includegraphics[trim={0cm 0cm 0cm 0cm},width=1.0\linewidth]{fig/methodv3.pdf}
%\caption{
%\emph{\ours: A \textsc{Wei}ghted \textsc{Com}posed Image Retrieval Method.} It utilizes a dual-encoder approach to process both \textcolor{imagequerycolor}{query image $y$} and \textcolor{textquerycolor}{query text $t$}. Initially, the \textcolor{imagequerycolor}{query image} is passed into a visual encoder $f$ and the \textcolor{textquerycolor}{query text} into a text encoder $g$, producing corresponding $d$-dimensional representations. Subsequently, similarity scores with the representations in the image dataset are calculated. These scores are then normalized and combined using a convex combination controlled by a $\lambda \in [0,1]$. Finally, an argmax(argsort) operation identifies the most relevant 
%and visually similar image, with attributes adjusted according to the text query, resulting in the 
%\textcolor{retrievedimagecolor}{retrieved image(s) $x$.}}
%\label{fig:method}
%\end{figure*}

\subsection{Problem formulation}
\label{sec:prelim}

In remote sensing composed image retrieval (RSCIR), the query consists of a reference image $y$ and a textual modifier $t$, denoted as $q=(y,t)$. The goal is to rank images $x \in X$ from a database according to a composed similarity score $s(q,x) \in \mathbb{R}$. Depending on the application, the query image $y$ is associated with either a semantic \emph{class} label $C_y$ or a \emph{scene} label $S_y$ identifying a geographic location. We denote by $A_y$ the attribute or state visible in $y$, and by $A_t$ the target attribute or state specified by the text.

We consider two relevance protocols:
\begin{itemize}[itemsep=2pt, parsep=0pt, topsep=2pt, leftmargin=1.2em]
    \item \textbf{Same class + target attribute.} A result $x$ is relevant if it depicts the same class as the query and matches the target attribute, i.e., $C_x=C_y$ and $A_x=A_t$. This setting captures attribute-oriented search, such as retrieving similar facilities with a different color, shape, density, or quantity.
    \item \textbf{Same scene + target state.} A result $x$ is relevant if it corresponds to the same scene/location as the query and matches the target state or change specified by the text, i.e., $S_x=S_y$ and $A_x=A_t$. This setting captures location-aware change-centric retrieval, such as retrieving the post-event image of the same site after a wildfire or flood.
\end{itemize}

The two protocols reflect different \emph{operational goals}. The first evaluates class-conditioned attribute modification, while the second evaluates identity-preserving state retrieval. The latter is not a minor variant of the former: preserving scene identity while satisfying a target state may favor different composition mechanisms.

To define $s$, we make use of pre-trained VLMs that consist of a \emph{visual encoder} \(f: \cI \to \mathbb{R}^d\) and a \emph{text encoder} \(g: \cT \to \mathbb{R}^d\), which map input images from image space \(\cI\) and words from the text space \(\cT\) to the same embedding space with dimension \(d\). We extract the visual embedding \(\mathbf{v}_y = f(y) \in \mathbb{R}^d\) and the text embedding \(\mathbf{v}_t = g(t) \in \mathbb{R}^d\) to use as queries. Finally, the embedding of a dataset image \(x \in X\) is denoted as \(\mathbf{v}_x = f(x) \in \mathbb{R}^d\). All embeddings are \(\ell_2\)-normalized.

\subsection{Baselines}
\label{sec:base}

\paragraph{Unimodal.}
These baselines score each database image using only one component of the composed query. Using the VLM encoders $f$ and $g$, we define the \emph{text-only} baseline as \( s_g(q,x) = g(t)^{\top} f(x) \) and the \emph{image-only} baseline as
\( s_f(q,x) = f(y)^{\top} f(x) \).
These baselines isolate the contribution of the textual modifier and visual reference, respectively.

\paragraph{Multimodal.} 
These baselines combine the two unimodal scores by fusing the image- and text-based similarities.
A simple fusion is \emph{averaging}: $s_a(q, x) = \frac{1}{2}[s_g(q, x) + s_f(q, x)]$, which yields the same ranking as \emph{summing} the similarities. We denote this baseline as \emph{text + image}.
While straightforward, this fusion can be biased toward the image signal in practice, since same-modality similarities (image--image) often have a different scale than cross-modal similarities (text--image), even under $\ell_2$ normalization.
A second fusion computes similarity by \emph{multiplying} the unimodal scores as $s_m(q, x) = s_f(q, x) \times s_g(q, x)$.
This method emphasizes retrieval results with high agreement between modalities, inherently penalizing disagreement, and serving as a form of \emph{soft normalization} that mitigates bias toward either modality. We denote this baseline as: \emph{text $\times$ image}.

\subsection{Methods}
\label{sec:methods}

Beyond unimodal and score-fusion baselines, we adapt and evaluate representative composed image retrieval methods from computer vision. These methods cover textual inversion, memory-based query construction, caption-and-LLM composition, diffusion-based embedding composition, supervised multimodal pooling, and training-free feature calibration.
%------------------------------------------------------------------------------
\paragraph{WeiCom}~\cite{psomas2024composedimageretrievalremote} is a training-free score-fusion method. It computes image-to-image and text-to-image similarities, standardizes each score distribution over the database, maps the standardized scores to $[0,1]$ using the Gaussian CDF, and combines them as $s_{\mathrm{WC}}(q,x)=\lambda s_g'(q,x)+(1-\lambda)s_f'(q,x)$, where $s_f'$ and $s_g'$ are the calibrated image and text scores, and $\lambda \in [0,1]$ controls the modality weight.
%------------------------------------------------------------------------------
\paragraph{Pic2Word}~\cite{pic2word} casts composition as \emph{textual inversion}: it learns an approximate inverse of the text encoder, mapping an image embedding back into a \emph{text-space token representation} that can be composed with the modifier. Concretely, given $\mathbf v_y=f(y)$, it predicts a pseudo-text representation $y^\star \approx g^{-1}(\mathbf v_y)$ with a small network trained on large-scale image--text pairs. The composed query is then formed in text space via concatenation and encoded as $g([y^\star; t])$, which is used for text-to-image retrieval.
%------------------------------------------------------------------------------
\paragraph{SEARLE}~\cite{searle} also follows textual inversion, but avoids pretraining by performing \emph{test-time optimization}. For each query image, it directly optimizes a pseudo token $y^\star$ such that $g(y^\star)$ matches $\mathbf v_y$ under a cosine objective, complemented by an LLM/GPT-based~\cite{gpt} regularizer that encourages linguistic plausibility. The optimized $y^\star$ is then concatenated with the modifier and encoded with $g(\cdot)$ for retrieval. This per-query optimization is computationally heavier, but can yield strong inversion quality.
%------------------------------------------------------------------------------
\paragraph{FreeDom}~\cite{freedom} replaces explicit inversion with a \emph{memory} of textual anchors. It pre-encodes a vocabulary $W=\{w_i\}_{i=1}^N$ through the text encoder $g(\cdot)$ to obtain $V_W=\{g(w_i)\}_{i=1}^N$ and performs nearest-neighbor search from the image embedding $\mathbf v_y$ to retrieve a set of words whose embeddings best match $\mathbf v_y$. Since the mapping between words and embeddings is explicit, this provides an interpretable textual surrogate for the image query without optimization or pretraining. The retrieved words are concatenated with the modifier and encoded with $g(\cdot)$ for text-to-image retrieval. FreeDom further improves robustness through query expansion using visually similar images.
%------------------------------------------------------------------------------
\paragraph{CIReVL}~\cite{cirevl} follows a \emph{caption-and-LLM} pipeline. It converts the image query to natural language via captioning rather than inversion. A captioner (\eg, BLIP/BLIP-2~\cite{blip, blip2}) produces an image description, which is then merged with the modifier using an LLM to form a refined query sentence. Retrieval is performed as text-to-image matching with the VLM. This pipeline favors human-readable intermediate representations and can benefit from strong captioning and rewriting quality.
%------------------------------------------------------------------------------
\paragraph{CompoDiff}~\cite{compodiff} recasts composition as a \emph{diffusion} process in the joint VLM embedding space. Starting from the image embedding $\mathbf v_y$, a denoising diffusion model gradually injects the textual condition $t$ (with image self-conditioning), producing a composed embedding $\tilde{\mathbf v}_{y,t}$ after $T$ steps. Retrieval is then performed with cosine similarity $\tilde{\mathbf v}_{y,t}^{\top}\mathbf v_x$.
%------------------------------------------------------------------------------
\paragraph{MagicLens}~\cite{zhang2024magiclensselfsupervisedimageretrieval} follows a \emph{supervised multimodal pooling} pipeline. It learns a dedicated composition module via supervision. It is trained on web-mined triplets (query image, query text, target image), and fine-tunes a VLM augmented with additional attention/pooling layers to map the joint image--text input to a single embedding. Database images are mapped to the same space and matched by cosine similarity.
%------------------------------------------------------------------------------
\paragraph{BASIC}~\cite{basic} is a \emph{training-free compositional scoring} method operating on frozen VLM features. It starts from the standard image-to-image and text-to-image similarities, and aims to make their combination more reliable by (i) reducing modality- and domain-specific bias, (ii) suppressing nuisance directions in the embedding space, and (iii) fusing the two modalities with an ``AND''-like criterion. Concretely, BASIC first \emph{centers} image and text embeddings by subtracting mean features (optionally estimated from a large external corpus), which mitigates global bias in the representation. It then applies a \emph{semantic projection} derived from two textual corpora: a positive/object corpus $C_{+}$ and a negative/style/context corpus $C_{-}$. A contrastive PCA-style construction emphasizes directions that are informative for object content while de-emphasizing directions associated with style, viewpoint, or acquisition context. Optionally, BASIC further strengthens the visual cue via \emph{query expansion} using top retrieved neighbors. Finally, BASIC computes modality scores against the database and applies a lightweight \emph{score calibration} to reduce scale mismatch between modalities, before combining them with a multiplicative fusion regularized by a Harris-like penalty.
%------------------------------------------------------------------------------
\section{Experiments}
\label{sec:exp}

\subsection{Experimental setup}
\label{sec:setup}

\paragraph{Datasets.}
We conduct our study on three datasets that cover both \emph{attribute-driven} and \emph{change-centric} composed retrieval in EO imagery. 
PatternCom~\cite{psomas2024composedimageretrievalremote} is an existing RSCIR benchmark derived from PatternNet~\cite{zhou2018patternnet}, a high resolution dataset with 38 classes (800 images per class, $256{\times}256$ pixels). PatternCom organizes PatternNet into composed queries of the form (query image, query text), where the text specifies a target \emph{attribute value} for the class depicted in the query image (\eg, shape: \texttt{oval} for swimming pool). It spans six attribute types (\emph{color, context, density, existence, quantity, shape}) with class-specific value sets and highly variable numbers of positives per query. In PatternCom, relevance is defined by the \emph{same class + target attribute} criterion. As illustrated in \autoref{fig:teaser} (a,b), this protocol naturally supports attribute-oriented RS use cases such as activity/occupancy monitoring and capacity/infrastructure monitoring.

\begin{figure*}[t]
    \centering
    \includegraphics[width=\textwidth]{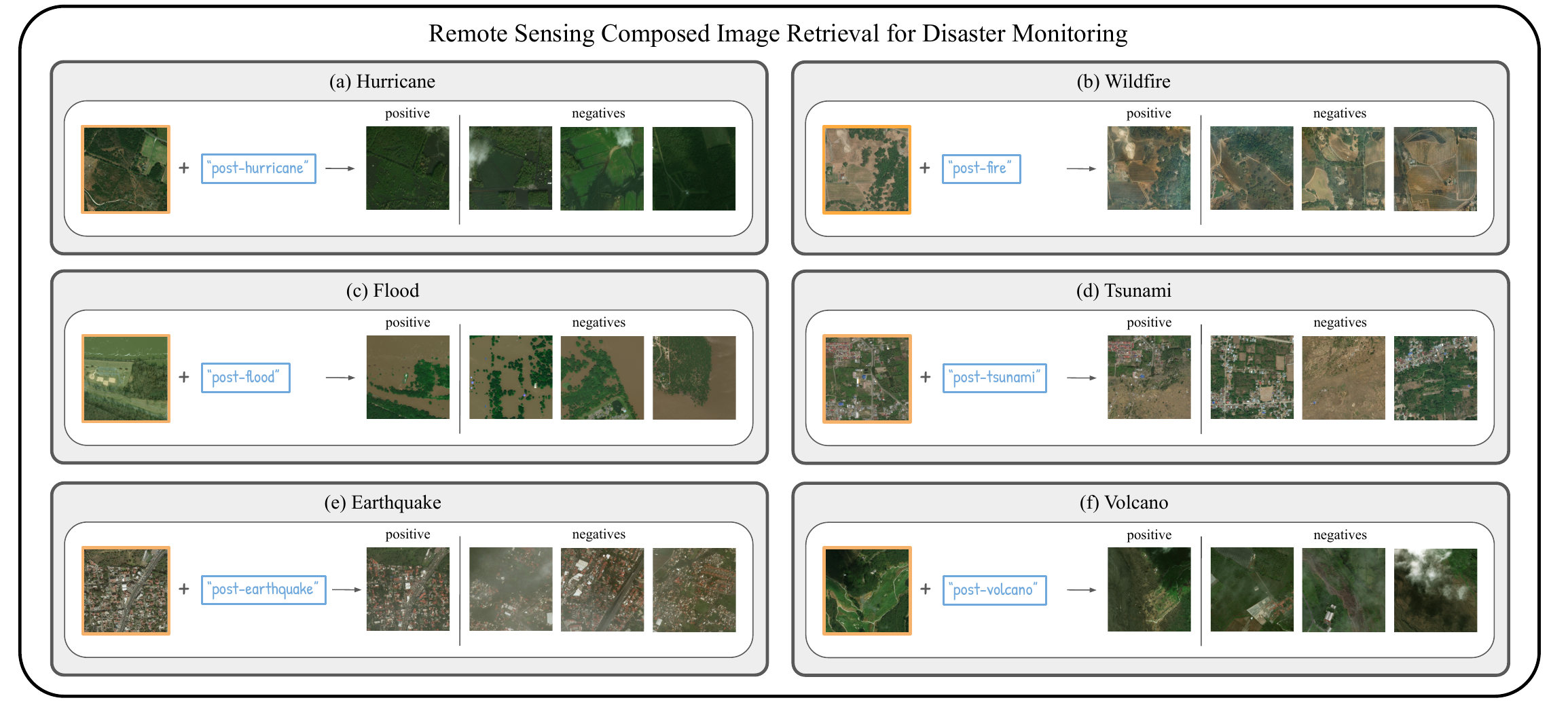}
    \vspace{-16pt}
    \caption{%
    \textbf{Remote sensing composed image retrieval for disaster monitoring.}
    We restructure xView2~\cite{xview} into a composed retrieval setting where a query consists of a \emph{pre-event} \textcolor{orange}{reference image} of a specific location and a \textcolor{textquerycolor}{textual modifier} describing the desired post-event state (\eg, \texttt{post-hurricane}). For each disaster type (a--f), we illustrate the query and the corresponding relevance criterion: \emph{positives} are images of the \emph{same scene/location} depicting the specified post-event state, while \emph{negatives} are non-matching candidates (\eg, different locations with the target change or the same location without the target change).%
    }
    \label{fig:xview}
    \vspace{-2mm}
\end{figure*}

PatternCom does not fully reflect operational scenarios where users seek \emph{the same location} under a different state. To bridge this gap, we introduce xView2-CIR, a new dataset built by restructuring xView2~\cite{xview} into a composed retrieval setting for disaster and damage monitoring. In xView2-CIR, each query pairs a \emph{pre-event} reference image of a geo-registered scene with a text modifier describing the desired \emph{post-event} state (\eg, \texttt{post-hurricane}, \texttt{post-fire}); relevance follows the \emph{same scene/location + target state/change} criterion, as depicted in \autoref{fig:xview}. The unique positive of each query is the post-event image of the same geo-registered location. All remaining images are treated as negatives, including images from different locations that exhibit the same disaster category and images from the same location that do not match the target post-event state, when applicable. This construction isolates the challenge of identity-preserving change retrieval.

Finally, to further highlight the applicability of RSCIR to urban change analysis, we include qualitative examples from LEVIR-CC~\cite{levir}, which provides paired imagery for building change detection. We use a small, partially structured subset to form composed queries for urban development (\eg, \texttt{new building}), following the \emph{same scene/location + target change} criterion; these examples are used for qualitative analysis only (\eg, \autoref{fig:teaser} (d)), and are not included in the quantitative benchmark due to the absence of a fully curated CIR protocol for this dataset in our study. A comprehensive overview of the dataset statistics is provided in~\ref{app:datasets}.

\paragraph{Networks.} We evaluate six vision-language models, all using the ViT-L/14 vision backbone: CLIP~\cite{openclip} LAION-2B~\cite{laion}, RemoteCLIP~\cite{liu2023remoteclip}, OpenAI CLIP~\cite{clip}, SigLIP~\cite{siglip}, CLIP LAION-RS~\cite{skyscript2023}, and SkyCLIP-50~\cite{skyscript2023}. We provide further details on the networks in~\ref{app:backbones}.

\paragraph{Vocabularies.} Recent CIR methods~\cite{searle, freedom, basic} leverage a vocabulary of possible categories or concepts. However, many of these, such as Open Images v7~\cite{open_images} with 21k concepts, originate from general-purpose computer vision datasets. To better align these methods with remote sensing semantics, we create a family of synthetic vocabularies using GPT-4o~\cite{gpt4o}, which we will distribute to avoid redistribution of any copyrighted content. These vocabularies, denoted as RSText, consist of increasing sizes: 150 to two thousand distinct concepts. The generation process (presented in~\autoref{app:voc}) follows a structured prompt that emphasizes fine-grained, diverse, and remote-sensing-relevant terminology across a wide range of thematic areas. To combine both general-purpose and domain-specific semantics, we further merge RS-Text-2k with the Open Images v7, resulting in a hybrid vocabulary of 23k classes, referred to as HybridText-23k.

\paragraph{Evaluation protocols.} 
For a query $q$, we compute Average Precision (AP) as the mean of the precision values at the ranks where relevant items are retrieved. We then aggregate AP across queries to obtain mAP, which reflects both relevance and ranking quality. For PatternCom, we follow an \emph{attribute-balanced} protocol: AP is first averaged over queries within each attribute type (\eg, \emph{color}), and the final score is the unweighted average across attribute types. This corresponds to a \emph{macro-averaged mAP} ($\mathrm{mAP}_{\text{macro}}$), ensuring that each attribute contributes equally. For xView2-CIR, the query distribution across disaster types is imbalanced. Therefore, we report both (i) the same \emph{macro-averaged} mAP over disaster categories, $\mathrm{mAP}_{\text{macro}}$, which weights each disaster type equally, and (ii) the standard \emph{overall} mAP, $\mathrm{mAP}_{\text{overall}}$, computed by averaging AP over \emph{all} queries irrespective of category. Reporting both provides a fair view of performance: $\mathrm{mAP}_{\text{macro}}$ reflects robustness across rare disaster types, while $\mathrm{mAP}_{\text{overall}}$ reflects expected performance under the natural query frequency.

\paragraph{Hyperparameters.} Both PatternCom and xView2-CIR are used strictly as evaluation sets. To avoid tuning on test data, we keep method hyperparameters fixed to the default settings reported in the original papers (or their official implementations) whenever applicable. When a method exposes several knobs (\eg, visual/textual expansion sizes, inversion iterations, diffusion steps), we do \emph{not} optimize them on PatternCom or xView2-CIR. Instead, we report a sensitivity analysis in the ablation section to characterize how performance varies with key hyperparameters, and we use a single fixed configuration for all main comparisons. For instance, for WeiCom we report the $\lambda$ sensitivity separately, while using a single fixed $\lambda$ in the main tables for fair comparison.

\subsection{Experimental analysis}
\label{sec:exp_analysis}

\paragraph{Benchmark.}~\autoref{tab:sota_1}, \ref{tab:sota_2}, and~\ref{tab:xview2_total} report the main quantitative results on PatternCom and xView2-CIR, respectively, covering a broad range of composition methods and vision--language models. We include unimodal baselines (\emph{text-only}, \emph{image-only}) and two simple multimodal fusions: score averaging (\emph{text $+$ image}) and score multiplication (\emph{text $\times$ image}). We further benchmark representative state-of-the-art CIR methods from the computer vision literature: CompoDiff~\cite{compodiff}, Pic2Word~\cite{pic2word}, SEARLE~\cite{searle},  CIReVL~\cite{cirevl}, MagicLens~\cite{zhang2024magiclensselfsupervisedimageretrieval}, WeiCom~\cite{psomas2024composedimageretrievalremote}, BASIC~\cite{basic}, and FreeDom~\cite{freedom}, after applying domain-grounded adaptations required for EO imagery. Our adaptation principle is intentionally conservative: for each method, we preserve the original inference mechanism and introduce only minimal domain-grounded modifications needed to make it meaningful for EO imagery, such as replacing generic vocabularies, adjusting prompt templates, or using remote-sensing-relevant textual resources. We generally avoid benchmark-specific tuning beyond what is required for compatibility. Detailed adaptation choices and implementation details are provided in~\ref{app:implementation}.

\subsection{Experimental results}
\label{sec:results}

%-----------------------------------------------
\begin{table}[!htbp]
\centering
\scriptsize
\caption{\textbf{Attribute modification performance on PatternCom (mAP, \%).} Results for three vision–-language backbones and up to eight composition methods. We report \emph{macro-averaged mAP} ($\mathrm{mAP}_{\text{macro}}$): for each attribute type, AP is averaged over all queries targeting a given attribute value (\eg, \texttt{rectangular} for Shape), then averaged over the remaining values of the same attribute type (\eg, \texttt{oval}, \texttt{kidney-shaped}), and finally averaged across attribute types. Avg.: resulting $\mathrm{mAP}_{\text{macro}}$ over all attribute types and classes; \best{bold}: best; \second{underline}: second.}
\label{tab:sota_1}
\par\medskip
\noindent\centering\textbf{(a) CLIP LAION-2B}\par
\centering
\scriptsize
\setlength{\tabcolsep}{2pt}
\begin{tabular}{lcccccc@{\hspace{4pt}}|@{\hspace{4pt}}c}
\toprule
\Th{Method} &
Color  &
Context &
Density    &
Existence   &
Quantity   &
Shape   &
Avg. \\ \midrule 
Text-only           & 13.47      & 4.83      & 3.58      & 2.00       & 3.31      & 6.22      & 5.57 \\
Image-only          & 14.66      & 8.32      & 13.49     & 16.47      & 7.84      & 15.76     & 12.74 \\
\midrule
Text $+$ Image   & 23.13      & 11.02     & 15.87     & 16.93      & 10.13      & 21.38 & 16.41  \\
Text $\times$ Image   & 40.97      & 11.87     & 14.48     & 14.36      & 20.31      & 23.99 & 21.00 \\
\midrule
%CompoDiff & 0.53 & 8.94 &	3.13 &	1.34 &	0.29 &	14.04 &	4.71 \\
%Pic2Word & 1.32	& 2.62 &	2.72 &	1.55 &	0.69 &	1.69 &	1.77 \\
SEARLE & 14.75 & 7.98 &	13.63 &	15.23 &	8.01 &	15.86 &	12.58 \\
CIReVL & 17.79 & \second{28.55} & \second{17.69} & \second{35.14} & 14.95 & \second{25.91} & \second{23.34} \\
%FreeDom   & 3.04      & 2.49     & 3.31     & 6.95      & 0.70 & 6.00 & 3.75 \\
%\midrule
WeiCom  & \second{46.08} & 17.45 & 16.49   & 8.36   & \second{18.15} & 23.97 & 21.75 \\
BASIC      & 29.26 & 6.38 & 15.29 & 18.54 & 12.18 & 17.18 & 16.47 \\
%\rowcolor{LighterSteel}
%WeiCom$_{.3}$ & 46.74 & 20.97 & 22.07   & 13.22   & 20.96 & 26.22 & 25.03 \\
%\midrule
%\oursnew & 46.55 &	43.49 & 21.32 &	46.98 &	25.09 &	48.13 &	38.59 \\

%FreeDom (500) & \textbf{46.85} &	\textbf{50.40} & \textbf{23.26} & \textbf{47.65} &	\textbf{27.55} &	\textbf{48.81} &	\textbf{40.75} \\
\rowcolor{appleteal!60}
FreeDom & \textbf{46.55} &	\textbf{43.49} & \textbf{21.32} & \textbf{46.98} &	\textbf{25.09} &	\textbf{48.13} &	\textbf{38.59} \\
\bottomrule
\end{tabular}

%\vspace{6pt}

\par\medskip
\noindent\centering\textbf{(b) RemoteCLIP}\par
\centering
\scriptsize
\setlength{\tabcolsep}{2pt}
\begin{tabular}{lcccccc@{\hspace{4pt}}|@{\hspace{4pt}}c}
\toprule
\Th{Method} &
Color  &
Context &
Density    &
Existence   &
Quantity   &
Shape   &
Avg. \\ \midrule 
Text-only            & 10.75    & 8.87          & 22.16       & 6.98     & 8.25  & 24.12  & 13.52        \\
Image-only           & 14.40    & 6.62          & 15.11       & 13.10     & 6.99   & 15.18  & 11.90       \\
\midrule
Text $+$ Image   & 23.67    & 10.01         & 18.45       & 13.98     & 7.97   & 19.63  & 15.62         \\
Text $\times$ Image & 47.20 & 19.65 & 27.09 & 12.97 & 14.59  & 40.60 & 27.02 \\
\midrule
CompoDiff & 8.58 & 17.88 & 14.41 & 6.04 & 9.34 & 12.46 & 11.45 \\
Pic2Word & 40.88 & \best{40.26} & 17.41 &	16.92 &	9.18 &	27.98 &	25.44 \\
SEARLE & 14.44 & 6.00 & 13.49 & 12.95 & 7.29 & 14.86 & 11.51 \\
CIReVL & 42.80 & 36.79 &	21.34	& \textbf{43.81} &	19.35 &	\second{35.55} &	\second{33.27} \\
%\midrule
WeiCom  & 43.68 & 31.45 & \best{39.94}   & 14.92   & 20.51    & 29.78    & 30.05 \\
BASIC      & \second{47.65} & 15.45 & 17.55 & 22.03 & \second{20.76} & 25.00 & 24.74 \\
%$\ours_{.6}$  & 41.04 & 31.59 & \textbf{41.56}   & 14.56   & 20.79   & 31.24   & 30.13 \\
%\midrule
%\oursnew & 49.8	& 38.8 &	28.31 &	36.64 &	26.77 &	39.49 &	36.64 \\
%\rowcolor{LightSteelBlue1}
%\oursnew & 49.8	& 38.8 &	28.31 &	36.64 &	26.77 &	39.49 &	36.64 \\
%FreeDom (500) & \textbf{51.10}	& \textbf{41.36} &	29.11 & 37.66 &	\textbf{29.48} &	\textbf{40.55} &	\textbf{38.32} \\
\rowcolor{appleteal!60}
FreeDom & \textbf{49.80}	& \underline{38.80} &	\second{28.31} & 36.64 &	\textbf{26.77} &	\textbf{39.49} &	\textbf{36.64} \\
\bottomrule
\end{tabular}

%\vspace{6pt}

\par\medskip
\noindent\centering\textbf{(c) OpenAI CLIP}\par
\centering
\scriptsize
\setlength{\tabcolsep}{2pt}
\begin{tabular}{lcccccc@{\hspace{4pt}}|@{\hspace{4pt}}c}
\toprule
\Th{Method} &
Color  &
Context &
Density    &
Existence   &
Quantity   &
Shape   &
Avg. \\ \midrule 
Text-only         & 3.49  & 7.75  & 2.51  & 3.31  & 1.01  & 2.94  & 3.50 \\
Image-only        & 13.61 & 7.86  & 18.19 & 13.30 & 8.30  & 15.08 & 12.72 \\
\midrule
Text $+$ Image & 16.59 & 11.13 & 19.88 & 15.13 & 8.96  & 17.55 & 14.87 \\
Text $\times$ Image & 27.59 & 14.65 & 15.63 & 17.97 & 9.08  & 17.77 & 17.12 \\
\midrule
CompoDiff & 10.70 & 8.32 & 16.83 & 12.36 & 4.67 & 13.50 & 11.06 \\
Pic2Word & 27.09 & 20.93 & 19.04   & 9.81   & 8.05 & 19.24 & 17.36 \\
SEARLE & 15.24 & 7.25 & 16.20 & 13.30 & 9.00 & 15.78 & 12.80 \\
CIReVL     & 29.32 & \second{31.14} & 16.02 & \second{37.23} & 11.70 & 22.88 & 24.72 \\
MagicLens  & 34.63 & 18.98 & 14.54 & 21.46 & 13.21 & 17.70 & 20.09 \\
%\midrule
WeiCom & 27.42 & 25.15 & 13.63 & 21.98 & 7.01  & 15.33 & 18.42 \\
%$\ours_{.2}$ & 31.76 & 22.74 & 29.16 & 21.96 & 9.43 & 17.60 & 22.11 \\
%\midrule
%\rowcolor{LightSteelBlue1}
% \oursnew & 39.08 & 45.87 &	36.72 &	37.25 &	13.22 &	28.69 &	33.47 \\
BASIC      & \second{38.31} & 24.53 & \second{26.35} & 30.94 & \best{18.03} & \best{35.73} & \second{28.98} \\
%FreeDom (500)    & \best{41.04} & \best{46.26} & \best{43.83} & \second{36.90} & 12.85 & \second{28.46} & \best{34.89} \\
\rowcolor{appleteal!60}
FreeDom    & \best{39.08} & \best{45.87} & \best{36.72} & \best{37.25} & \second{13.22} & \second{28.69} & \best{33.47} \\
\bottomrule
\end{tabular}
\end{table}
%-----------------------------------------------
%-----------------------------------------------
\begin{table}[!htbp]
\centering
\scriptsize
\caption{\textbf{Attribute modification performance on PatternCom (mAP, \%).} Results for three vision–-language backbones and up to seven composition methods. We report \emph{macro-averaged mAP} ($\mathrm{mAP}_{\text{macro}}$): for each attribute type, AP is averaged over all queries targeting a given attribute value (\eg, \texttt{rectangular} for Shape), then averaged over the remaining values of the same attribute type (\eg, \texttt{oval}, \texttt{kidney-shaped}), and finally averaged across attribute types. Avg.: resulting $\mathrm{mAP}_{\text{macro}}$ over all attribute types and classes; \best{bold}: best; \second{underline}: second.}
\label{tab:sota_2}
\par\medskip
\noindent\centering\textbf{(a) SigLIP}\par
\centering
\scriptsize
\setlength{\tabcolsep}{2pt}
\begin{tabular}{lcccccc@{\hspace{4pt}}|@{\hspace{4pt}}c}
\toprule
\Th{Method} &
Color  &
Context &
Density    &
Existence   &
Quantity   &
Shape   &
Avg. \\ \midrule 
Text-only              & 7.99  & 3.93  & 2.30  & 9.93  & 0.58  & 31.16 & 9.32 \\
Image-only             & 15.03 & 10.67 & 21.46 & 21.13 & 8.94  & 17.37 & 15.77 \\
\midrule
Text $+$ Image   & 25.11 & 11.46 & 20.47 & 21.31 & 10.66 & 27.92 & 19.49 \\
Text $\times$ Image  & 18.80 & 5.96  & 6.57  & 18.77 & 1.28  & \second{43.66} & 15.84 \\
\midrule
WeiCom & 24.77 & \second{15.32} & 9.06  & 22.37 & 4.59  & 43.64 & 19.96 \\
BASIC      & \second{36.11} & 10.30 & \second{21.76} & \second{24.68} & \second{20.43} & 19.46 & \second{22.12} \\
%$\ours_{.1}$ & 37.10 & 14.66 & 22.66 & 25.91 & 14.03 & 30.95 & 24.22 \\
%\midrule
%\rowcolor{LightSteelBlue1}
% \oursnew & 47.99	& 49.88	& 22.84	& 54.7 &	37.05 &	67.17 &	46.61 \\
%FreeDom (500) & \textbf{48.58} &	\textbf{51.68} &	\textbf{24.10} &	\textbf{54.20} &	\textbf{41.83} &	\textbf{68.92} &	\textbf{48.22} \\
\rowcolor{appleteal!60}
FreeDom & \textbf{47.99} &	\textbf{49.88} &	\textbf{22.84} &	\textbf{54.70} &	\textbf{37.05} &	\textbf{67.17} &	\textbf{46.61} \\
\bottomrule
\end{tabular}

%\vspace{6pt}

\par\medskip
\noindent\centering\textbf{(b) CLIP LAION-RS}\par
\centering
\scriptsize
\setlength{\tabcolsep}{2pt}
\begin{tabular}{lcccccc@{\hspace{4pt}}|@{\hspace{4pt}}c}
\toprule
\Th{Method} &
Color  &
Context &
Density    &
Existence   &
Quantity   &
Shape   &
Avg. \\ \midrule 
Text-only       & 3.85       & 6.13      & 2.60      & 5.56       & 0.99      & 4.57      & 3.95 \\
Image-only      & 13.35      & 7.51      & 15.82     & 15.79      & 7.50      & 15.91     & 12.65 \\
\midrule
Text $+$ Image  & 20.94      & 10.07     & 16.54     & 16.91      & 8.75      & 20.78     & 15.67 \\
Text $\times$ Image & 21.38      & 12.98     & 10.91     & 16.91      & 10.20     & 22.32     & 15.78 \\
\midrule
CompoDiff & 19.70 & 11.32 & \second{17.98} & 17.97 & 9.67 & 20.36 & 16.17 \\
Pic2Word & 31.18 &	19.58 &	16.53 &	11.87 &	7.88 &	19.03 &	17.68 \\
SEARLE & 14.89 & 10.23 & 14.34 & 17.20 & 8.31 & 16.51 & 13.58 \\
CIReVL  & 32.57 & \second{36.45} &	15.33 &	\second{34.68} &	\textbf{13.78} &	24.88 &	\second{26.28} \\
%\midrule
WeiCom   & 29.63      & 30.45     & 11.19     & 21.13      & 8.60      & \second{25.85}     & 21.14 \\
BASIC      & \second{35.51} & 14.06 & 16.32 & 25.56 & 12.18 & 21.81 & 20.91 \\
%$\ours_{.3}$   & 33.42 & 30.81 & 18.22 & 23.22 & 10.17 & 27.38 & 23.87 \\
%\midrule
%\rowcolor{LightSteelBlue1}
% \oursnew & 41.78 &	51.34 &	21.88 &	38.22 &	13.29 &	33.58 &	33.35 \\
%FreeDom (500) & \textbf{41.86}	& \textbf{51.49} &	\textbf{26.12} &	\textbf{37.78} &	13.06 &	\textbf{34.11} &	\textbf{34.07} \\
\rowcolor{appleteal!60}
FreeDom & \textbf{41.78}	& \textbf{51.34} &	\textbf{21.88} &	\textbf{38.22} &	\second{13.29} &	\textbf{33.58} &	\textbf{33.35} \\
\bottomrule
\end{tabular}

%\vspace{6pt}

\par\medskip
\noindent\centering\textbf{(c) SkyCLIP-50}\par
\centering
\scriptsize
\setlength{\tabcolsep}{2pt}
\begin{tabular}{lcccccc@{\hspace{4pt}}|@{\hspace{4pt}}c}
\toprule
\Th{Method} &
Color  &
Context &
Density    &
Existence   &
Quantity   &
Shape   &
Avg. \\ \midrule 
Text-only       & 4.60       & 8.44      & 3.44      & 5.61       & 1.02      & 7.34      & 5.08 \\
Image-only      & 14.48      & 9.03      & 17.82     & 20.11      & 8.77      & 16.32     & 14.42 \\
\midrule
Text $+$ Image   & 23.49      & 12.64     & \second{19.54}     & 21.42      & 10.24     & 21.80     & 18.19 \\
Text $\times$ Image  & 37.03      & 15.84     & 16.24     & 23.99      & 12.90     & 31.51     & 22.92 \\
\midrule
CompoDiff & 20.59 & 12.67 & 17.57 & 21.98 & 10.23 & 22.31 & 17.56 \\
Pic2Word &	34.57 &	21.58 &	18.23 &	13.09 &	10.03 &	21.93 &	19.91 \\
SEARLE & 16.57 & 11.68 & 16.41 & 20.22 & 9.22 & 16.88 & 15.16 \\
CIReVL	& 34.90 &	34.93 &	17.09 &	\second{35.03} &	\second{14.39} &	28.92 &	\second{27.54} \\
%\midrule
WeiCom   & \second{40.46}      & \second{38.10}     & 18.17     & 27.91      & 10.10     & \second{31.52}     & 27.71 \\
BASIC      & 34.89 & 12.20 & 16.64 & 28.98 & 12.32 & 18.30 & 20.56 \\
%$\ours_{.3}$   & 44.39 & 37.65 & 27.16 & 32.39 & 12.43 & 30.88 & 30.82 \\
%\midrule
%\rowcolor{LightSteelBlue1}
% \oursnew & 49.88 & 50.47 &	29.71 &	38.00 &	14.58 &	39.99 &	37.11 \\
%FreeDom (500) & \textbf{51.65} &	\textbf{50.93} &	\textbf{33.60} &	\textbf{37.25} &	14.16 &	\textbf{40.98} &	\textbf{38.99} \\
\rowcolor{appleteal!60}
FreeDom & \textbf{49.88} &	\textbf{50.47} &	\textbf{29.71} &	\textbf{38.00} &	\best{14.58} &	\textbf{39.99} &	\textbf{37.11} \\
\bottomrule
\end{tabular}
\end{table}
%-----------------------------------------------

\paragraph{Quantitative results on PatternCom} 
\autoref{tab:sota_1} and~\ref{tab:sota_2} summarize attribute-modification retrieval performance on PatternCom across six VLM backbones. Unimodal baselines are consistently weak: \emph{image-only} usually outperforms \emph{text-only}, confirming that visual similarity alone is useful but insufficient for satisfying the textual modifier. Simple multimodal fusion provides clear gains over unimodal retrieval. In particular, \emph{text $\times$ image} is often stronger than \emph{text $+$ image}, although it is less stable on some backbones, such as SigLIP.

Among composition methods, FreeDom is the strongest performer, achieving the best average mAP across all six backbones, with particularly high absolute performance on SigLIP (46.61\%) and strong results on RemoteCLIP, SkyCLIP-50, and CLIP LAION-RS. CIReVL and BASIC form the next tier. CIReVL performs particularly well on remote-sensing-adapted CLIP backbones like RemoteCLIP and SkyCLIP-50. BASIC is the second strongest with OpenAI CLIP and SigLIP. WeiCom remains a competitive lightweight baseline, while MagicLens shows that supervised multimodal heads trained on natural-image triplets can transfer to EO imagery to some extent. In contrast, continuous inversion and diffusion-based approaches are generally less competitive. Pic2Word and SEARLE yield modest gains at best, while CompoDiff is often unstable and can even underperform the \emph{image-only} baseline. This suggests that methods relying on synthesized query embeddings or pseudo-token inversion are more sensitive to domain shift in EO imagery.

Overall, PatternCom favors methods that preserve class identity while injecting attribute-specific textual cues. Strong textual surrogates and domain-grounded vocabularies are therefore particularly beneficial, explaining the consistent advantage of FreeDom in this setting.

%-----------------------------------------------
\begin{table}[!htbp]
\scriptsize
\centering
\caption{\textbf{Disaster monitoring performance on xView2-CIR (mAP, \%).} Results for two vision--language backbones and three composition methods.
We report per-disaster mAP (\%) for \texttt{post-*} modifiers (\eg, Hurricane).
Avg.: $\mathrm{mAP}_{\text{macro}}$ across disasters (equal weight per disaster); Total: $\mathrm{mAP}_{\text{overall}}$ (weighted by the number of queries per disaster);
\best{bold}: best; \second{underline} second.}
\label{tab:xview2_total}
\vspace{-4pt}
%------------------------------------------------------------------------------
\par\medskip
\noindent\centering\textbf{(a) OpenAI CLIP}\par
\centering
\scriptsize
\setlength{\tabcolsep}{1pt}
\resizebox{0.48\textwidth}{!}{
\begin{tabular}{lcccccc@{\hspace{4pt}}|@{\hspace{4pt}}c@{\hspace{4pt}}|@{\hspace{4pt}}c}
\toprule
\Th{Method} &
Hurricane  &
Wildfire &
Flood    &
Tsunami   &
Earthquake   &
Volcano   &
Avg. &
Total \\ \midrule
Text-only           &  2.50 &  1.61 &  3.25 & 18.08 &  4.19 & 15.77 &  7.57 &  2.97 \\
Image-only          &  6.99 &  2.39 &  1.44 & 25.17 &  8.35 &  9.69 &  9.00 &  5.53 \\
\midrule
Text $+$ Image      & 14.61 &  7.45 &  4.75 & 37.46 & 16.63 & 25.49 & 17.73 & 12.14 \\
Text $\times$ Image & 18.11 & 12.12 &  8.36 & 37.63 & \second{17.93} & \best{61.46} & 25.94 & 16.35 \\
\midrule
WeiCom              & \best{25.94} & 13.88 & \second{9.43} & \second{42.96} & \best{36.48} & 55.34 & \best{30.67} & \best{21.40} \\
BASIC               & \second{22.71} & \best{18.49} & \best{14.59} & \best{46.15} &  1.69 & \second{59.20} & \second{27.14} & \second{21.38} \\
%$\ours_{.2}$        & 27.72 & 11.98 &  9.05 & 49.70 & 39.02 & 52.86 & 31.72 & 21.84 \\
FreeDom             & 18.01 & \second{17.46} &  3.77 & 35.03 &  3.21 & 25.88 & 17.23 & 16.84 \\
\bottomrule
\end{tabular}
}

%\vspace{6pt}

\par\medskip
\noindent\centering\textbf{(b) SigLIP}\par
\centering
\scriptsize
\setlength{\tabcolsep}{1pt}
\resizebox{0.48\textwidth}{!}{
\begin{tabular}{lcccccc@{\hspace{4pt}}|@{\hspace{4pt}}c@{\hspace{4pt}}|@{\hspace{4pt}}c}
\toprule
\Th{Method} &
Hurricane  &
Wildfire &
Flood    &
Tsunami   &
Earthquake   &
Volcano   &
Avg. &
Total \\ \midrule 
Text-only           &  2.32 &  1.81 &  1.09 &  7.10 &  0.88 & 13.77 &  4.49 &  2.31 \\
Image-only          & 10.65 &  9.74 & 10.68 & 24.15 & 14.44 & 20.77 & 15.07 & 10.97 \\
\midrule
Text $+$ Image      & 15.86 & 19.94 & 13.05 & 26.29 & \second{19.92} & 26.17 & 20.20 & 17.50 \\
Text $\times$ Image & 10.74 & 13.69 & \second{14.20} & 17.54 & \best{23.17} & \second{53.96} & 22.22 & 13.08 \\
\midrule
WeiCom              & \second{20.19} & \best{30.61} &  9.95 & \second{29.46} & 18.56 & 32.09 & \second{23.48} & \best{23.14} \\
BASIC               & \best{21.54} & 12.30 & 13.98 & \best{36.52} & 13.86 & \best{56.79} & \best{25.83} & \second{18.52} \\
%$\ours_{.2}$        & 26.05 & 27.30 & 15.06 & 31.50 & 33.06 & 40.97 & 29.68 & 27.30 \\
FreeDom             & 13.07 & \second{15.37} & \best{21.22} & 19.33 &  1.60 & 48.61 & 19.99 & 15.37 \\
\bottomrule
\end{tabular}
}
%------------------------------------------------------------------------------
\vspace{-6pt}
\end{table}
%-----------------------------------------------

\paragraph{Quantitative results on xView2-CIR} 
\autoref{tab:xview2_total} reports performance on xView2-CIR under the \emph{same scene/location + target state} criterion. Unimodal baselines remain weak, while score fusion provides clear gains across both backbones. \emph{Text $+$ image} is consistently reliable, whereas \emph{text $\times$ image} is more variable across disaster types. Among composition methods, WeiCom is the most robust overall, achieving the best \Th{Total} score for both OpenAI CLIP and SigLIP, suggesting that calibrated modality fusion is well suited to identity-preserving change retrieval. In contrast, FreeDom and BASIC are less effective than on PatternCom, partly because their query-expansion mechanisms can introduce visually similar but geographically different scenes, which is harmful when relevance depends on scene identity. These results show that change-centric RSCIR poses different challenges from attribute-based retrieval and should be evaluated as a distinct setting.

%-----------------------------------------------
\begin{figure*}[!htbp]
\scriptsize
\centering
\setlength{\tabcolsep}{1.2pt}
\newcommand{\sz}{1.12cm}
\makebox[\linewidth]{%
\begin{tabular}{c@{\hspace{-8pt}}c@{\hspace{8pt}}cc@{\hspace{4pt}}|@{\hspace{4pt}}c@{\hspace{8pt}}cc@{\hspace{4pt}}|@{\hspace{4pt}}c@{\hspace{8pt}}cc}
\toprule
& Query & 1 & 2 & Query & 1 & 2 & Query & 1 & 2 \\
\midrule

% =========================
% (a) Image-only
% =========================
\multirow{2}{*}{\makecell[c]{(a)\\Image-only}\hspace{2em}}
& \qcell{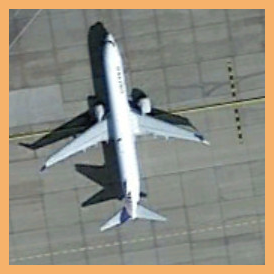}{purple}
& \includegraphics[width=\sz,height=\sz]{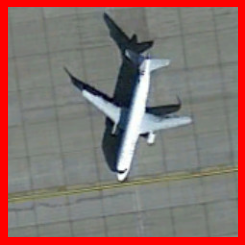}
& \includegraphics[width=\sz,height=\sz]{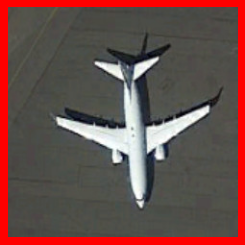}
& \qcell{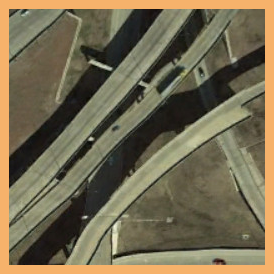}{water}
& \includegraphics[width=\sz,height=\sz]{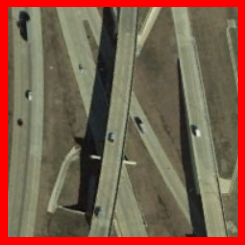}
& \includegraphics[width=\sz,height=\sz]{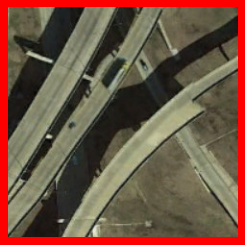}
& \qcell{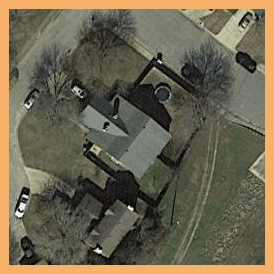}{dense}
& \includegraphics[width=\sz,height=\sz]{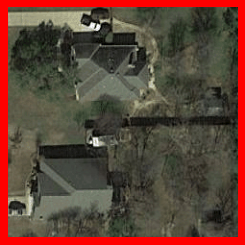}
& \includegraphics[width=\sz,height=\sz]{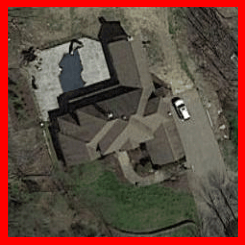} \\
\addlinespace[4pt]
& \qcell{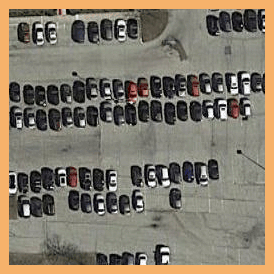}{empty}
& \includegraphics[width=\sz,height=\sz]{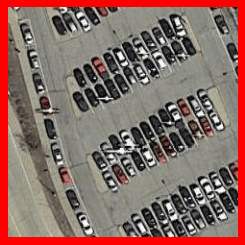}
& \includegraphics[width=\sz,height=\sz]{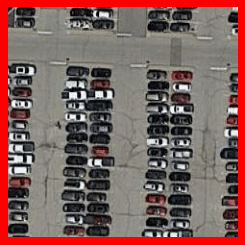}
& \qcell{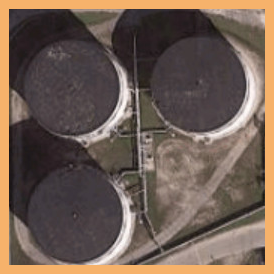}{one}
& \includegraphics[width=\sz,height=\sz]{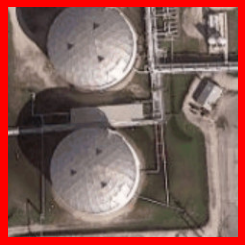}
& \includegraphics[width=\sz,height=\sz]{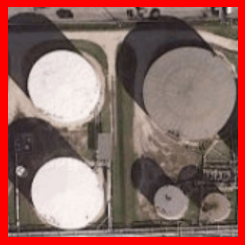}
& \qcell{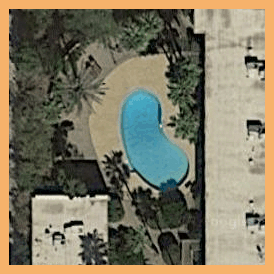}{rectan.}
& \includegraphics[width=\sz,height=\sz]{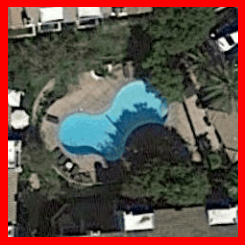}
& \includegraphics[width=\sz,height=\sz]{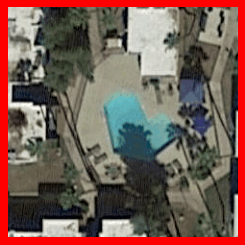} \\

\midrule

% =========================
% (b) Text-only
% =========================
\multirow{2}{*}{\makecell[c]{(b)\\Text-only}\hspace{2em}}
& \qcell{fig/retrieval/queries/airplane743.png}{purple}
& \includegraphics[width=\sz,height=\sz]{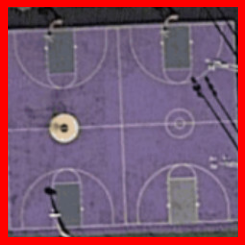}
& \includegraphics[width=\sz,height=\sz]{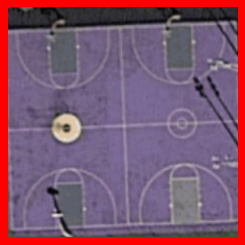}
& \qcell{fig/retrieval/queries/overpass353.png}{water}
& \includegraphics[width=\sz,height=\sz]{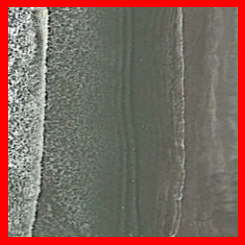}
& \includegraphics[width=\sz,height=\sz]{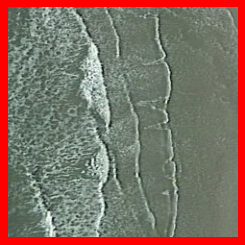}
& \qcell{fig/retrieval/queries/sparseresidential245.png}{dense}
& \includegraphics[width=\sz,height=\sz]{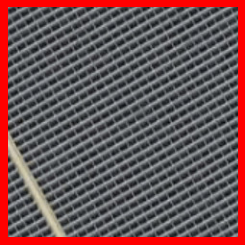}
& \includegraphics[width=\sz,height=\sz]{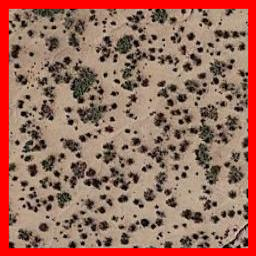} \\
\addlinespace[4pt]
& \qcell{fig/retrieval/queries/parkinglot373.png}{empty}
& \includegraphics[width=\sz,height=\sz]{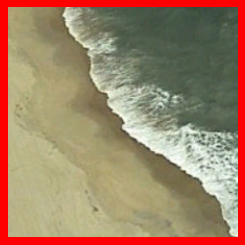}
& \includegraphics[width=\sz,height=\sz]{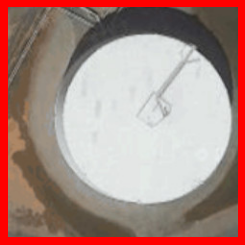}
& \qcell{fig/retrieval/queries/storagetank244.png}{one}
& \includegraphics[width=\sz,height=\sz]{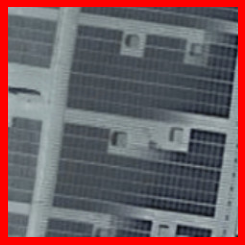}
& \includegraphics[width=\sz,height=\sz]{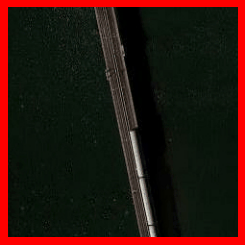}
& \qcell{fig/retrieval/queries/swimmingpool270.png}{rectan.}
& \includegraphics[width=\sz,height=\sz]{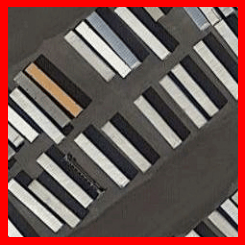}
& \includegraphics[width=\sz,height=\sz]{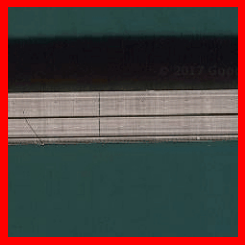} \\

\midrule

% =========================
% (c) WeiCom
% =========================
\multirow{2}{*}{\makecell[c]{(c)\\WeiCom}\hspace{2em}}
& \qcell{fig/retrieval/queries/airplane743.png}{purple}
& \includegraphics[width=\sz,height=\sz]{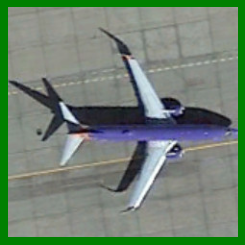}
& \includegraphics[width=\sz,height=\sz]{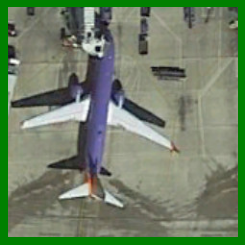}
& \qcell{fig/retrieval/queries/overpass353.png}{water}
& \includegraphics[width=\sz,height=\sz]{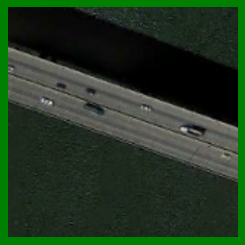}
& \includegraphics[width=\sz,height=\sz]{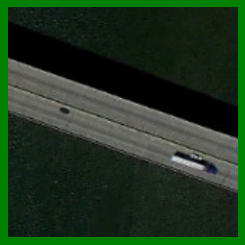}
& \qcell{fig/retrieval/queries/sparseresidential245.png}{dense}
& \includegraphics[width=\sz,height=\sz]{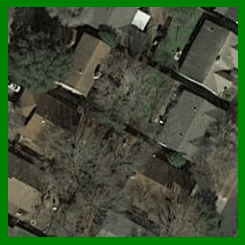}
& \includegraphics[width=\sz,height=\sz]{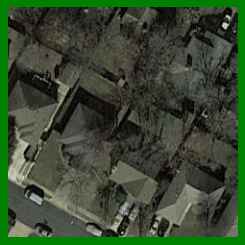} \\
\addlinespace[4pt]
& \qcell{fig/retrieval/queries/parkinglot373.png}{empty}
& \includegraphics[width=\sz,height=\sz]{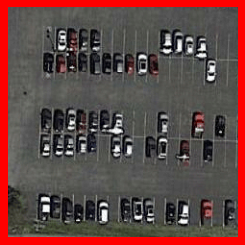}
& \includegraphics[width=\sz,height=\sz]{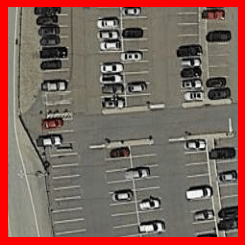}
& \qcell{fig/retrieval/queries/storagetank244.png}{one}
& \includegraphics[width=\sz,height=\sz]{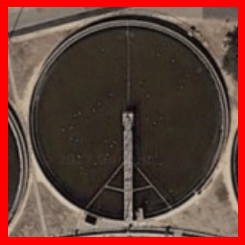}
& \includegraphics[width=\sz,height=\sz]{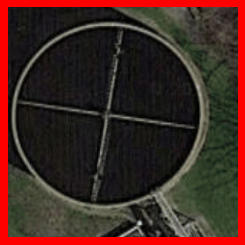}
& \qcell{fig/retrieval/queries/swimmingpool270.png}{rectan.}
& \includegraphics[width=\sz,height=\sz]{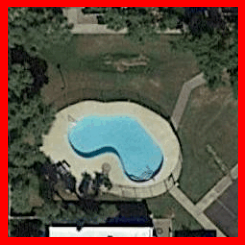}
& \includegraphics[width=\sz,height=\sz]{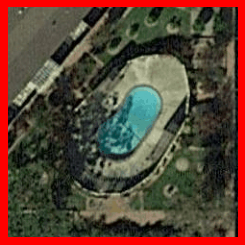} \\

\midrule

% =========================
% (d) FreeDom
% =========================
\multirow{2}{*}{\makecell[c]{(d)\\FreeDom}\hspace{2em}}
& \qcell{fig/retrieval/queries/airplane743.png}{purple}
& \includegraphics[width=\sz,height=\sz]{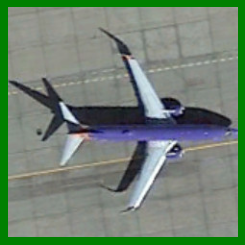}
& \includegraphics[width=\sz,height=\sz]{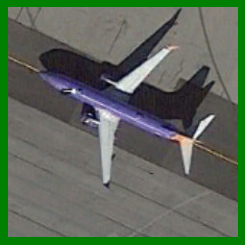}
& \qcell{fig/retrieval/queries/overpass353.png}{water}
& \includegraphics[width=\sz,height=\sz]{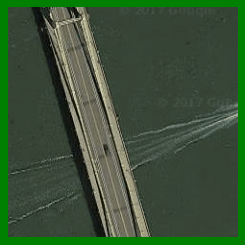}
& \includegraphics[width=\sz,height=\sz]{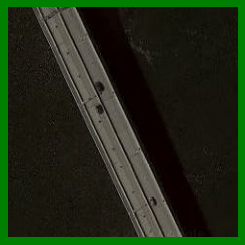}
& \qcell{fig/retrieval/queries/sparseresidential245.png}{dense}
& \includegraphics[width=\sz,height=\sz]{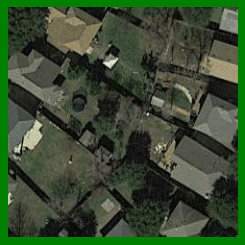}
& \includegraphics[width=\sz,height=\sz]{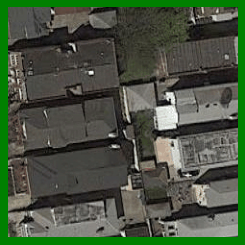} \\
\addlinespace[4pt]
& \qcell{fig/retrieval/queries/parkinglot373.png}{empty}
& \includegraphics[width=\sz,height=\sz]{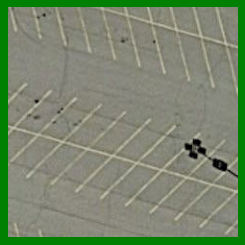}
& \includegraphics[width=\sz,height=\sz]{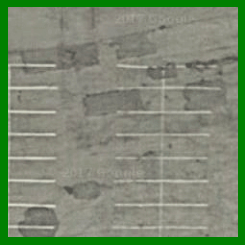}
& \qcell{fig/retrieval/queries/storagetank244.png}{one}
& \includegraphics[width=\sz,height=\sz]{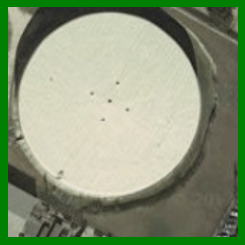}
& \includegraphics[width=\sz,height=\sz]{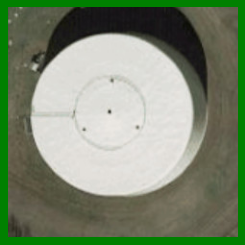}
& \qcell{fig/retrieval/queries/swimmingpool270.png}{rectan.}
& \includegraphics[width=\sz,height=\sz]{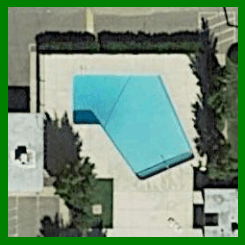}
& \includegraphics[width=\sz,height=\sz]{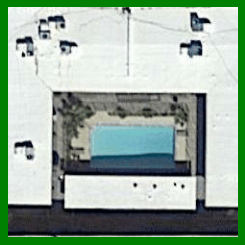} \\

\bottomrule
\end{tabular}%
}

\caption{\textbf{Qualitative composed retrieval results on PatternCom} with OpenAI CLIP. Comparison between unimodal and multimodal methods. Each query is shown as a \textcolor{orange}{reference image} combined with a boxed \textcolor{textquerycolor}{textual modifier}. Columns report the top-2 retrieved results. Retrieval is evaluated under the \emph{same class + target attribute value} relevance criterion.}
\label{fig:retrievals}
\end{figure*}
%-----------------------------------------------
\paragraph{Qualitative results on PatternCom.} 
In~\autoref{fig:retrievals}, we visualize representative retrieval outputs on PatternCom with OpenAI CLIP, comparing unimodal baselines (image-only, text-only) against two multimodal methods (WeiCom and FreeDom). The examples clearly illustrate the limitations of unimodal retrieval. WeiCom generally improves over unimodal baselines by leveraging both modalities, but it can still be sensitive to modality dominance depending on the attribute type. In the \emph{activity monitoring} example (parking lot + \texttt{empty}), the retrieved results remain biased toward the visual layout of the reference image, yielding parking-lot-like structures that do not clearly satisfy the “empty” intent. In the \emph{capacity monitoring} example (storage tanks + \texttt{one}), the textual cue can dominate, pulling results with fewer tanks that are only loosely aligned with the specific query instance. In contrast, FreeDom is consistently the most reliable across all shown cases, producing results that simultaneously respect the query class and the target attribute value, aligning well with PatternCom’s relevance criterion.
%-----------------------------------------------

%-----------------------------------------------
\begin{figure*}[!htbp]
    \centering
    \includegraphics[width=0.8\textwidth]{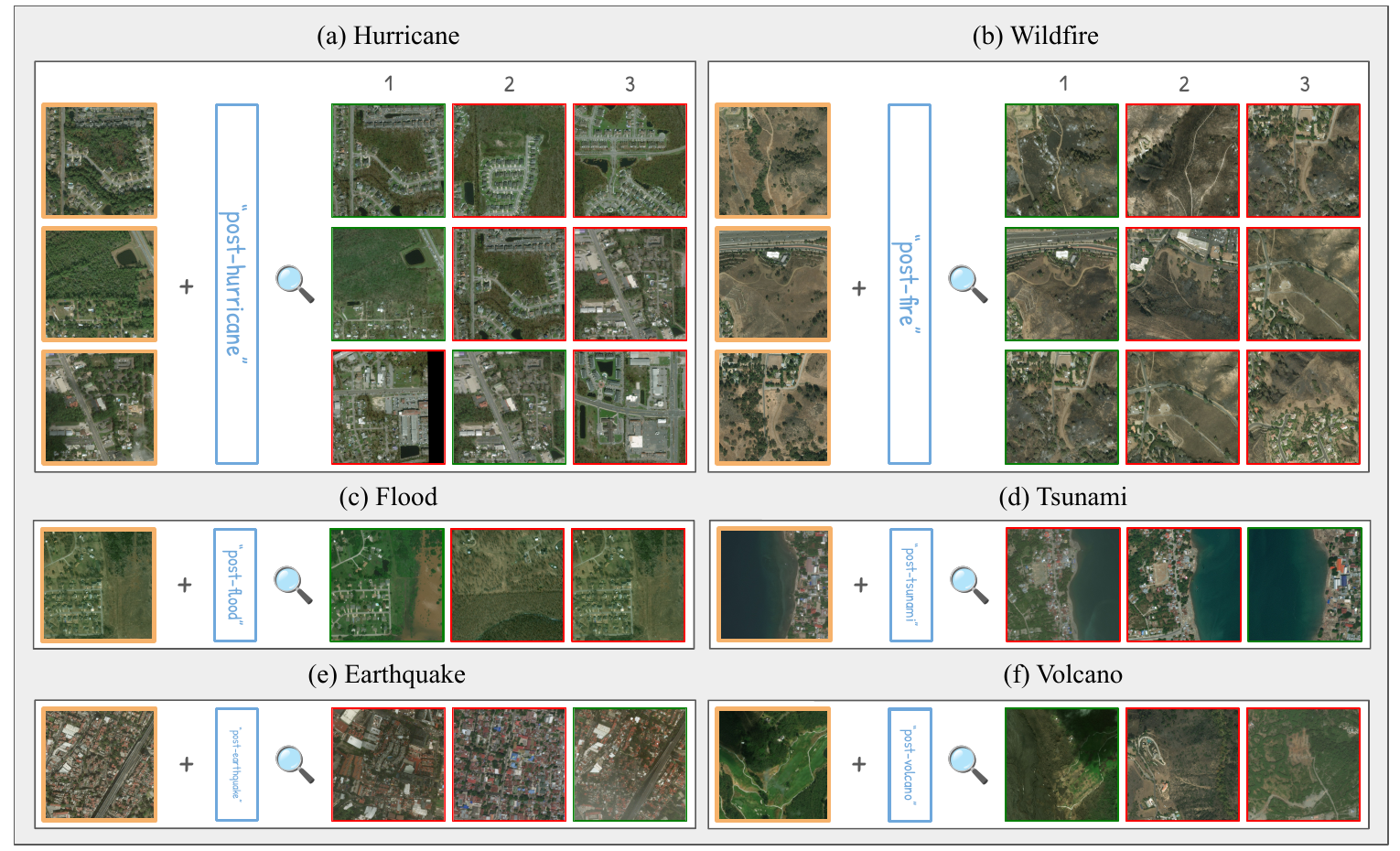}
    \caption{\textbf{Qualitative composed retrieval results on xView2-CIR} with OpenAI CLIP and WeiCom. Each query combines a \emph{pre-disaster} \textcolor{orange}{reference image} with a boxed \textcolor{textquerycolor}{textual modifier} (\texttt{post-*}) indicating the target post-event state. We show the top retrieved \emph{post-disaster} results per disaster type. Retrieval is evaluated under the \emph{same scene/location + target state} relevance criterion, so visually plausible post-event matches from different locations are considered negatives.}
    \label{fig:xview_retrievals}
    \vspace{-2mm}
\end{figure*}
%-----------------------------------------------
%-----------------------------------------------
\begin{figure*}[!htbp]
    \centering
    \includegraphics[width=\textwidth]{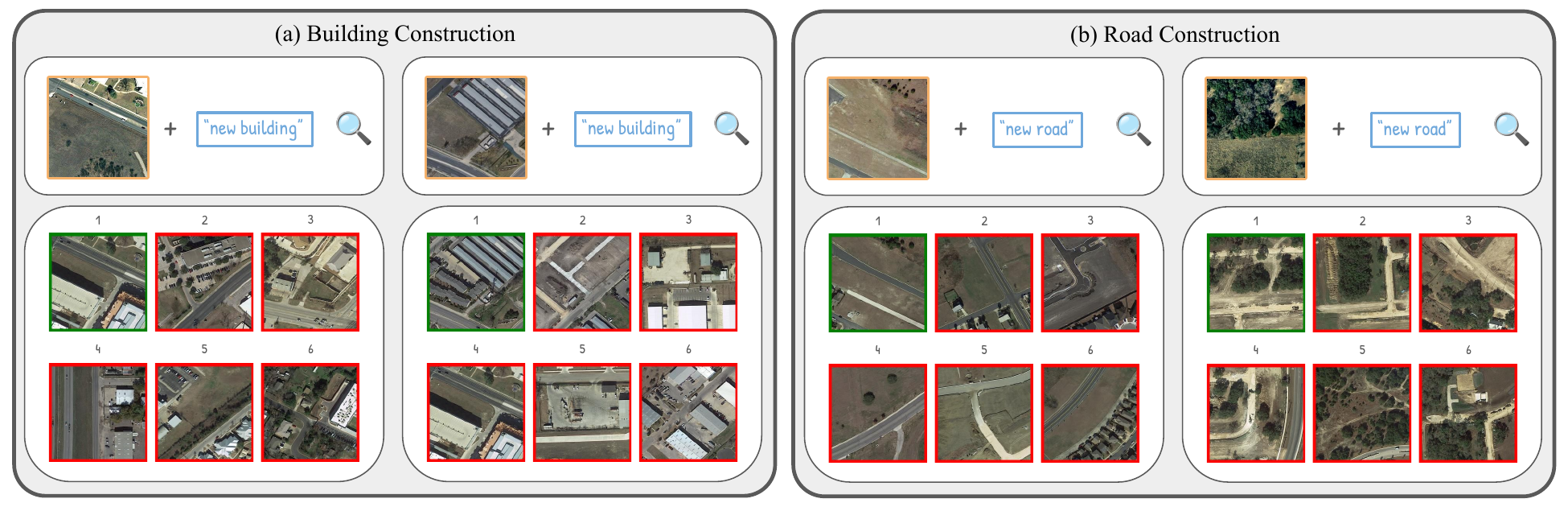}
    \vspace{-12pt}
    \caption{\textbf{Qualitative composed retrieval on LEVIR-CC} with OpenAI CLIP and WeiCom. Each query is shown as a \textcolor{orange}{reference image} combined with a \textcolor{textquerycolor}{textual change modifier}. We report the top retrieved candidates under the \emph{same location + target state} relevance criterion.}
    \label{fig:levir_retrievals}
    \vspace{-2mm}
\end{figure*}
%-----------------------------------------------

\paragraph{Qualitative results on xView2-CIR.} In~\autoref{fig:xview_retrievals}, we visualize qualitative composed retrieval results on xView2-CIR using OpenAI CLIP with WeiCom. Unlike PatternCom, relevance in xView2-CIR requires \emph{the same scene or location} \emph{and} the \emph{target state} (post-event), which exposes a key failure mode of generic composition: many retrieved images match the requested disaster state but drift to a different geographic area with similar visual cues. This is especially apparent for hurricanes and wildfires, where widespread debris/burn patterns and textured backgrounds can look plausible across locations, leading to several off-scene false positives.

Still, WeiCom succeeds in a subset of cases where the scene has distinctive geometry or landmarks (\eg, road layouts, coastline structure, dense urban blocks), allowing the query image to anchor location while the text steers the retrieval toward the correct post-event state. Overall, the figure highlights that xView2-CIR is less forgiving than attribute-only benchmarks: effective composed retrieval must jointly preserve \emph{instance identity} (same place) while applying a \emph{state change} constraint, and methods that lean too heavily on “state-like” appearance cues tend to retrieve the right disaster but the wrong scene. 
%-----------------------------------------------

\paragraph{Qualitative results on LEVIR-CC.}
To illustrate more applied EO use cases,~\autoref{fig:levir_retrievals} shows qualitative results on LEVIR-CC~\cite{levir} using WeiCom with OpenAI CLIP. As shown in the examples for \emph{building construction} and \emph{road construction}, the method takes a reference image together with a change-oriented modifier (\eg, \texttt{new building}) and retrieves candidate scenes that match the requested change, under the \emph{same location/scene + target state} relevance criterion. In these cases, the top retrieved results frequently correspond to plausible construction changes, suggesting that composed retrieval can serve as a lightweight interface for change-focused search in real remote sensing workflows (\eg, rapid infrastructure monitoring).
%-----------------------------------------------
\subsection{Sensitivity and Ablation Analysis}
\label{sec:ablation}

\paragraph{Impact of $\lambda$ in WeiCom.}
\autoref{fig:ablation_weicom_lambda_patterncom} analyzes the modality-control parameter $\lambda$ of WeiCom with RemoteCLIP, where $\lambda{=}0$ corresponds to \emph{image-only} and $\lambda{=}1$ to \emph{text-only} retrieval. Although all main results use a fixed $\lambda{=}0.5$ to avoid tuning on PatternCom, the sensitivity analysis shows that performance is maximized at intermediate values, with the best average mAP obtained at $\lambda{=}0.6$. Similar trends are observed across backbones, where the optimal $\lambda$ consistently lies away from the unimodal extremes (e.g., $\lambda{=}0.3$ for CLIP LAION-2B). This broad mid-range optimum indicates that both the visual reference and textual modifier contribute meaningfully to attribute-based retrieval. The main exception is \emph{color}, which peaks at smaller $\lambda$ values, suggesting stronger dependence on the visual signal.

\begin{casfigurehere}
\centering
\scriptsize
\begin{tikzpicture}
\begin{axis}[
    width=7cm,
    height=3.85cm,
    xlabel={$\lambda$},
    ylabel={mAP (\%)},
    xmin=0, xmax=1,
    xtick={0,0.1,0.2,0.3,0.4,0.5,0.6,0.7,0.8,0.9,1.0},
    xmajorgrids=true,
    ymajorgrids=true,
    grid style={gray!10},
    legend style={
      at={(1.02,0.99)},
      anchor=north west,
      draw=black,
      fill=white,
      fill opacity=0.9,
      text opacity=1,
      font=\tiny
    },
    legend cell align=right,
    every axis plot/.append style={line width=1.2pt},
    mark=none,
    cycle list={
        {blue!55},{orange!65},{ForestGreen!55},{red!55},{purple!55},{teal!55},{black!55}
    }
]

\addplot
coordinates {(0,14.5) (0.1,55.3) (0.2,53.0) (0.3,49.6) (0.4,46.4) (0.5,43.7) (0.6,41.0) (0.7,38.2) (0.8,35.0) (0.9,30.4) (1.0,10.8)};
\addlegendentry{Color}

\addplot coordinates {(0,6.6) (0.1,13.3) (0.2,20.2) (0.3,25.7) (0.4,29.5) (0.5,31.5) (0.6,31.6) (0.7,29.6) (0.8,24.8) (0.9,16.9) (1.0,8.9)};
\addlegendentry{Context}

\addplot coordinates {(0,15.1) (0.1,23.3) (0.2,29.5) (0.3,34.0) (0.4,37.4) (0.5,39.9) (0.6,41.6) (0.7,42.0) (0.8,40.7) (0.9,35.9) (1.0,22.2)};
\addlegendentry{Density}

\addplot coordinates {(0,13.1) (0.1,14.2) (0.2,14.5) (0.3,15.0) (0.4,15.0) (0.5,14.9) (0.6,14.6) (0.7,13.8) (0.8,12.5) (0.9,10.3) (1.0,7.0)};
\addlegendentry{Existence}

\addplot coordinates {(0,7.0) (0.1,17.6) (0.2,18.9) (0.3,19.7) (0.4,20.2) (0.5,20.5) (0.6,20.8) (0.7,20.9) (0.8,20.8) (0.9,20.1) (1.0,8.3)};
\addlegendentry{Quantity}

\addplot coordinates {(0,15.2) (0.1,23.8) (0.2,24.7) (0.3,26.2) (0.4,28.0) (0.5,29.8) (0.6,31.2) (0.7,32.0) (0.8,32.0) (0.9,31.3) (1.0,24.1)};
\addlegendentry{Shape}

\addplot coordinates {(0,11.9) (0.1,24.6) (0.2,26.8) (0.3,28.4) (0.4,29.4) (0.5,30.0) (0.6,30.1) (0.7,29.4) (0.8,27.6) (0.9,24.2) (1.0,13.6)};
\addlegendentry{Average}

\end{axis}
\end{tikzpicture}
\vspace{-6pt}
\casfigcaption{fig:ablation_weicom_lambda_patterncom}{\textbf{Impact of modality control $\lambda$} in WeiCom with RemoteCLIP on PatternCom. Curves report attribute-wise mAP (\%) as $\lambda$ varies.}
%\label{fig:ablation_weicom_lambda_patterncom}
\end{casfigurehere}
%-----------------------------------------------

\paragraph{Impact of vocabulary on FreeDom.} 
\autoref{fig:vocab_freedom_patterncom} analyzes how the textual memory affects FreeDom on PatternCom with CLIP LAION-2B. Open Images v7 provides strong performance due to its large and diverse vocabulary of roughly 21k concepts. However, the EO-specific RSText vocabularies remain competitive despite being much smaller, with up to 200$\times$ fewer entries for RSText-150 and 20$\times$ fewer entries for RSText-1k. This suggests that compact, domain-grounded vocabularies can provide effective textual surrogates for EO imagery. The best performance is obtained by HybridText-23k, which combines Open Images with RSText-2k, indicating that broad semantic coverage and EO-specific grounding are complementary.

\begin{casfigurehere}
\centering
\scriptsize
\begin{adjustbox}{max width=0.9\linewidth}
\begin{tikzpicture}
\begin{axis}[
    ybar,
    bar width=5pt,
    width=11cm,
    height=5.5cm,
    enlargelimits=0.15,
    ylabel={mAP (\%)},
    symbolic x coords={
        OIv7, RST-150, RST-1k, HT-23k
    },
    xtick=data,
    xticklabels={
    {\scriptsize OIv7}, 
    {\scriptsize RST-150}, 
    {\scriptsize RST-1k}, 
    {\scriptsize HT-23k}
},
    xmajorgrids=true,
    ymajorgrids=true,
    grid style={gray!10},
    ymin=0, ymax=55,
    legend style={at={(0.5,1.1)}, anchor=south, legend columns=7},
    legend cell align={left},
    nodes near coords,
    every node near coord/.append style={font=\tiny, rotate=90, anchor=west, color=black}
]

% Attribute: Color
\addplot+[fill=blue!55, draw=blue!55] coordinates {
    (OIv7, 44.11) (RST-150, 13.48)
    (RST-1k, 34.25) (HT-23k, 46.55)
};

% Attribute: Context
\addplot+[fill=orange!65, draw=orange!65] coordinates {
    (OIv7, 47.85) (RST-150, 12.12) 
    (RST-1k, 43.17) (HT-23k, 43.49)
};

% Attribute: Density
\addplot+[fill=ForestGreen!55, draw=ForestGreen!55] coordinates {
    (OIv7, 15.38) (RST-150, 5.90) 
    (RST-1k, 17.93) (HT-23k, 21.32)
};

% Attribute: Existence
\addplot+[fill=red!55, draw=red!55] coordinates {
    (OIv7, 45.68) (RST-150, 36.59) 
    (RST-1k, 40.61)  (HT-23k, 46.98)
};

% Attribute: Quantity
\addplot+[fill=purple!55, draw=purple!55] coordinates {
    (OIv7, 24.91) (RST-150, 20.55) 
    (RST-1k, 21.26) (HT-23k, 25.09)
};

% Attribute: Shape
\addplot+[fill=teal!55, draw=teal!55] coordinates {
    (OIv7, 45.07) (RST-150, 37.08) 
    (RST-1k, 34.43) (HT-23k, 48.13)
};

% Attribute: Average
\addplot+[fill=black!55, draw=black!55] coordinates {
    (OIv7, 37.17) (RST-150, 20.95)
    (RST-1k, 31.94) (HT-23k, 38.59)
};

\legend{
    Color, Context, Density, Existence, Quantity, Shape, Average
}

\end{axis}
\end{tikzpicture}
\end{adjustbox}
\vspace{-4pt}
\casfigcaption{fig:vocab_freedom_patterncom}{\textbf{Impact of vocabulary} on FreeDom with CLIP LAION-2B on PatternCom. We compare Open Images v7, EO-specific RSText vocabularies, and their hybrid combination HybridText-23k. Bars report mAP (\%) per attribute and the average.}
%\label{fig:vocab_freedom_patterncom}
\end{casfigurehere}
%-----------------------------------------------
\paragraph{Impact of textual expansion on FreeDom.} 
\autoref{fig:freedom_freq_expansion} analyzes frequency-based textual expansion in FreeDom by disabling visual query expansion ($k{=}1$) and varying the number of aggregated textual concepts ($m{=}n$). On PatternCom, increasing $m{=}n$ improves average mAP from 28.99\% at $m{=}n{=}1$ to 35.89\% at $m{=}n{=}7$, with the largest gains on shape and context. The same moderate-expansion trend appears on xView2-CIR, where performance improves from 9.59\% to 16.84\% at $m{=}n{=}7$. In both settings, larger expansions plateau or degrade, suggesting that overly long textual surrogates introduce noisy or off-target concepts. The corresponding xView2-CIR analysis is reported in~\ref{app:additional_ablations}.

\begin{casfigurehere}
\centering
\begin{adjustbox}{max width=\linewidth}
\begin{tikzpicture}
\begin{axis}[
    width=9cm,
    height=4.5cm,
    ylabel={mAP (\%)},
    xlabel={Hyperparameters ($k{=}1$, varying $m{=}n$)},
    xtick=data,
    symbolic x coords={k1_mn1, k1_mn3, k1_mn5, k1_mn7, k1_mn10, k1_mn15},
    xticklabels={
        $1$,
        $3$,
        $5$,
        $7$,
        $10$,
        $15$
    },
    xticklabel style={anchor=north},
    ymin=10, ymax=50,
    xmajorgrids=true,
    ymajorgrids=true,
    grid style={gray!10},
    legend style={
      at={(1.02,0.99)},
      anchor=north west,
      draw=black,
      fill=white,
      fill opacity=0.9,
      text opacity=1,
      font=\scriptsize
    },
    legend cell align={left},
    %every axis plot/.append style={line width=1.2pt, solid}
    %every node near coord/.append style={font=\tiny},
    %nodes near coords,
    tick label style={font=\scriptsize},
    label style={font=\scriptsize},
]

% Attribute-wise plots
\addplot+[mark=*, thick, line width=1.2pt, color=blue!55, mark options={fill=blue!55}] coordinates {
    (k1_mn1, 41.41) (k1_mn3, 43.9) (k1_mn5, 44.07)
    (k1_mn7, 44.63) (k1_mn10, 44.1) (k1_mn15, 43.18)
};
\addlegendentry{Color}

\addplot+[mark=*, thick, line width=1.2pt,color=orange!65, mark options={fill=orange!65}] coordinates {
    (k1_mn1, 28.8) (k1_mn3, 38.07) (k1_mn5, 39.61)
    (k1_mn7, 40.48) (k1_mn10, 43.1) (k1_mn15, 46.96)
};
\addlegendentry{Context}

\addplot+[mark=*, thick, line width=1.2pt,color=ForestGreen!55, mark options={fill=ForestGreen!55}] coordinates {
    (k1_mn1, 13.7) (k1_mn3, 18.84) (k1_mn5, 19.69)
    (k1_mn7, 19.07) (k1_mn10, 16.86) (k1_mn15, 13.69)
};
\addlegendentry{Density}

\addplot+[mark=*, thick, line width=1.2pt,color=red!55, mark options={fill=red!55}] coordinates {
    (k1_mn1, 43.65) (k1_mn3, 42.71) (k1_mn5, 43.22)
    (k1_mn7, 44.45) (k1_mn10, 43.57) (k1_mn15, 41.29)
};
\addlegendentry{Existence}

\addplot+[mark=*, thick, line width=1.2pt,color=purple!55, mark options={fill=purple!55}] coordinates {
    (k1_mn1, 20.00) (k1_mn3, 22.83) (k1_mn5, 23.15)
    (k1_mn7, 22.54) (k1_mn10, 21.99) (k1_mn15, 21.70)
};
\addlegendentry{Quantity}

\addplot+[mark=*, thick, solid, line width=1.2pt,color=teal!55, mark options={fill=teal!55}] coordinates {
    (k1_mn1, 26.42) (k1_mn3, 34.4) (k1_mn5, 40.95)
    (k1_mn7, 44.14) (k1_mn10, 44.2) (k1_mn15, 41.87)
};
\addlegendentry{Shape}

\addplot+[mark=*, thick, solid, line width=1.2pt,color=black!55, mark options={fill=black!55}] coordinates {
    (k1_mn1, 28.99) (k1_mn3, 33.46) (k1_mn5, 35.12)
    (k1_mn7, 35.89) (k1_mn10, 35.64) (k1_mn15, 34.78)
};
\addlegendentry{Average}

\end{axis}
\end{tikzpicture}
\end{adjustbox}
\vspace{-16pt}
\casfigcaption{fig:freedom_freq_expansion}{\textbf{Impact of textual expansion} on FreeDom with CLIP LAION-2B on PatternCom. Visual expansion is disabled ($k{=}1$), and the number of aggregated textual concepts is varied as $m{=}n$.}
%\label{fig:freedom_freq_expansion}
\end{casfigurehere}
%-----------------------------------------------

\paragraph{Impact of visual expansion on FreeDom.}
\autoref{fig:freedom_visual_expansion} and~\ref{fig:freedom_visual_expansion_xview2} analyze visual query expansion in FreeDom by varying the number of proxy images $k$ while disabling textual expansion ($m{=}n{=}1$). On PatternCom, visual expansion is \emph{beneficial} up to a moderate range: average mAP increases from 28.99\% at $k{=}1$ to 30.41\% at $k{=}100$, before dropping to 26.71\% at $k{=}1000$. This is consistent with the \emph{same class + target attribute} criterion, where nearest neighbors often remain class-consistent and provide useful attribute evidence, until overly large proxy sets introduce noisy distractors. In contrast, visual expansion is \emph{harmful} on xView2-CIR: Total mAP decreases from 9.59\% at $k{=}1$ to 4.51\% at $k{=}500$. Since xView2-CIR requires the \emph{same scene/location + target state}, visually similar proxies usually correspond to different locations and inject irrelevant content. This explains why proxy-based expansion benefits attribute-oriented retrieval but can degrade identity-preserving change retrieval.

\begin{casfigurehere}
\centering
\begin{adjustbox}{max width=\linewidth}
\begin{tikzpicture}
\begin{axis}[
    width=9cm,
    height=4.5cm,
    ylabel={mAP (\%)},
    xlabel={Number of proxy images $k$ (with $m{=}n{=}1$)},
    symbolic x coords={k1, k10, k50, k100, k200, k500, k1000},
    xticklabels={1, 10, 50, 100, 200, 500, 1000},
    xtick=data,
    xticklabel style={anchor=north},
    ymin=5, ymax=50,
    xmajorgrids=true,
    ymajorgrids=true,
    grid style={gray!10},
    legend style={
      at={(1.02,0.99)},
      anchor=north west,
      draw=black,
      fill=white,
      fill opacity=0.9,
      text opacity=1,
      font=\scriptsize
    },
    legend cell align={left},
    tick label style={font=\scriptsize},
    label style={font=\scriptsize},
]

\addplot+[mark=*, thick, color=blue!55, mark options={fill=blue!55}] coordinates {
    (k1, 41.41) (k10, 43.18) (k50, 43.09)
    (k100, 42.71) (k200, 42.46) (k500, 41.14) (k1000, 39.09)
};
\addlegendentry{Color}

\addplot+[mark=*, thick, line width=1.2pt,color=orange!55, mark options={fill=orange!55}] coordinates {
    (k1, 28.8) (k10, 28.49) (k50, 27.88)
    (k100, 28.43) (k200, 28.75) (k500, 23.2) (k1000, 19.83)
};
\addlegendentry{Context}

\addplot+[mark=*, thick, line width=1.2pt,color=ForestGreen!55, mark options={fill=ForestGreen!55}] coordinates {
    (k1, 13.7) (k10, 12.46) (k50, 11.01)
    (k100, 10.73) (k200, 9.35) (k500, 8.16) (k1000, 7.3)
};
\addlegendentry{Density}

\addplot+[mark=*, thick, line width=1.2pt,color=red!55, mark options={fill=red!55}] coordinates {
    (k1, 43.65) (k10, 45.49) (k50, 46.38)
    (k100, 46.44) (k200, 46.56) (k500, 47.72) (k1000, 47.18)
};
\addlegendentry{Existence}

\addplot+[mark=*, thick, line width=1.2pt,color=purple!55, mark options={fill=purple!55}] coordinates {
    (k1, 20.00) (k10, 22.02) (k50, 24.37)
    (k100, 25.56) (k200, 26.69) (k500, 28.7) (k1000, 27.65)
};
\addlegendentry{Quantity}

\addplot+[mark=*, thick, line width=1.2pt,solid, color=teal!55, mark options={fill=teal!55}] coordinates {
    (k1, 26.42) (k10, 26.97) (k50, 28.35)
    (k100, 28.6) (k200, 28.32) (k500, 28.03) (k1000, 19.23)
};
\addlegendentry{Shape}

\addplot+[mark=*, thick, line width=1.2pt,solid, color=black!55, mark options={fill=black!55}] coordinates {
    (k1, 28.997) (k10, 29.768) (k50, 30.18)
    (k100, 30.412) (k200, 30.355) (k500, 29.492) (k1000, 26.713)
};
\addlegendentry{Average}

\end{axis}
\end{tikzpicture}
\end{adjustbox}
\vspace{-16pt}
\casfigcaption{fig:freedom_visual_expansion}{\textbf{Impact of visual expansion} on FreeDom with CLIP LAION-2B on PatternCom. The number of proxy images $k$ is varied while textual expansion is disabled ($m{=}n{=}1$).}
%\label{fig:freedom_visual_expansion}
\end{casfigurehere}

\begin{casfigurehere}
\centering
\begin{adjustbox}{max width=\linewidth}
\begin{tikzpicture}
\begin{axis}[
    width=9cm,
    height=4.5cm,
    ylabel={mAP (\%)},
    xlabel={Number of proxy images $k$ (with $m{=}n{=}1$)},
    symbolic x coords={k1, k5, k10, k20, k50, k100, k200, k500},
    xticklabels={1, 5, 10, 20, 50, 100, 200, 500},
    xtick=data,
    xticklabel style={anchor=north},
    ymin=2.6, ymax=10.5,
    xmajorgrids=true,
    ymajorgrids=true,
    grid style={gray!10},
    legend style={
      at={(1.02,0.99)},
      anchor=north west,
      draw=black,
      fill=white,
      fill opacity=0.9,
      text opacity=1,
      font=\scriptsize
    },
    legend cell align={left},
    tick label style={font=\scriptsize},
    label style={font=\scriptsize},
]

\addplot+[mark=*, thick, line width=1.2pt,color=blue!55, mark options={fill=blue!55}] coordinates {
    (k1, 9.93) (k5, 8.68) (k10, 7.54) (k20, 7.72) (k50, 7.02)
    (k100, 3.89) (k200, 4.75) (k500, 3.03)
};
\addlegendentry{Hurricane}

\addplot+[mark=*, thick, line width=1.2pt,color=orange!55, mark options={fill=orange!55}] coordinates {
    (k1, 8.46) (k5, 7.85) (k10, 5.04) (k20, 4.72) (k50, 4.26)
    (k100, 4.25) (k200, 4.16) (k500, 3.85)
};
\addlegendentry{Wildfire}

\addplot+[mark=*, thick, line width=1.2pt,color=ForestGreen!55, mark options={fill=ForestGreen!55}] coordinates {
    (k1, 5.41) (k5, 5.39) (k10, 4.88) (k20, 3.67) (k50, 3.46)
    (k100, 3.04) (k200, 3.14) (k500, 2.95)
};
\addlegendentry{Flood}

%\addplot+[mark=*, thick, line width=1.2pt, color=red!55, mark options={fill=red!55}] coordinates {
%    (k1, 34.01) (k5, 34.01) (k10, 34.01) (k20, 26.60) (k50, 24.94)
%    (k100, 24.94) (k200, 25.61) (k500, 32.79)
%};
%\addlegendentry{Tsunami}

%\addplot+[mark=*, thick, line width=1.2pt,color=purple!55, mark options={fill=purple!55}] coordinates {
%    (k1, 0.95) (k5, 0.95) (k10, 1.08) (k20, 1.26) (k50, 1.10)
%    (k100, 1.04) (k200, 1.04) (k500, 1.24)
%};
%\addlegendentry{Earthquake}

%\addplot+[mark=*, thick, line width=1.2pt,solid, color=teal!55, mark options={fill=teal!55}] coordinates {
%    (k1, 10.08) (k5, 10.19) (k10, 10.19) (k20, 6.14) (k50, 2.02)
%    (k100, 26.46) (k200, 30.59) (k500, 26.45)
%};
%\addlegendentry{Volcano}

\addplot+[mark=*, thick, line width=1.2pt, solid, color=black!55, mark options={fill=black!55}] coordinates {
    (k1, 9.59) (k5, 8.76) (k10, 7.18) (k20, 6.77) (k50, 6.13)
    (k100, 4.84) (k200, 5.33) (k500, 4.51)
};
\addlegendentry{Total}

\end{axis}
\end{tikzpicture}
\end{adjustbox}
\vspace{-16pt}
\casfigcaption{fig:freedom_visual_expansion_xview2}{\textbf{Impact of visual expansion} on FreeDom with OpenAI CLIP on xView2-CIR. The number of proxy images $k$ is varied while textual expansion is disabled ($m{=}n{=}1$).}
%\label{fig:freedom_visual_expansion_xview2}
\end{casfigurehere}
%-----------------------------------------------

\paragraph{Ablation of BASIC components.}
We further analyze the contribution of BASIC components, with full results reported in~\ref{app:additional_ablations}. On PatternCom, the largest drops occur when removing \emph{centering} or \emph{semantic projection}, indicating that modality calibration and semantic subspace alignment are the main drivers of BASIC's performance. Harris weighting, min-based normalization, text contextualization, and query expansion provide smaller, secondary gains. A similar trend appears on xView2-CIR, where centering and semantic projection remain the most important components, while visual query expansion becomes harmful under the \emph{same scene/location + target state} criterion. This supports the broader finding that proxy-based expansion can help attribute-oriented retrieval but may degrade identity-preserving change retrieval.
%-----------------------------------------------

\paragraph{Sensitivity to text query phrasing.}
We also evaluate whether composed retrieval is sensitive to the phrasing of the textual modifier, with full results reported in~\ref{app:additional_ablations}. On PatternCom, the original plain attribute values (\eg, \texttt{blue}, \texttt{water}, \texttt{dense}) are generally the most reliable, especially for stronger multimodal methods. More verbose templates such as \texttt{having blue color} or \texttt{with the main object modified to have blue color} often reduce performance, suggesting that additional function words can dilute the compact attribute signal in this benchmark.

On xView2-CIR, prompt sensitivity is more dataset- and method-dependent. Rephrasing \texttt{post-\{disaster\}} into impact-oriented or disaster-explicit descriptions (\eg, \texttt{burned area}, \texttt{flooded area}, \texttt{fire-affected region}) can benefit methods that rely more strongly on semantic text structure, particularly BASIC. Overall, these results indicate that prompt phrasing is an important practical factor: compact modifiers are preferable for controlled attribute edits, whereas change-centric retrieval may benefit from more descriptive event semantics.

%\paragraph{Joint hyperparameter search on $k$, $m$, and $n$ in \oursnew}
%In~\autoref{tab:freedom_final_ablation}, we present a comprehensive ablation study exploring the effect of jointly tuning the number of proxy images $k$, the number of retrieved labels per proxy $n$, and the number of merged concepts $m$ in \oursnew. Initially, we fix $n{=}1$ and search over different combinations of $k \in \{10, 50, 100, 200, 500, 1000\}$ and $m \in \{1, 3, 7, 15\}$. As shown in \autoref{tab:freedom_final_ablation} (a), performance consistently improves with higher values of $m$, and the optimal configuration under $n{=}1$ is observed at $k{=}500$, $m{=}7$ (35.68\% mAP).
%Next, fixing this optimal configuration of $k{=}500$, $m{=}7$, we sweep over increasing values of $n \in \{1, 2, 3, 5, 7, 9\}$. Results in \autoref{tab:freedom_final_ablation} (b) show that increasing $n$ continues to boost performance, reaching a peak of 40.75\% mAP at $n{=}7$. This trend is consistently verified across multiple values of $k$, confirming that deeper visual-textual alignment through diverse label expansion improves compositional retrieval. We adopt $k{=}500$, $m{=}7$, and $n{=}7$ as the final configuration for \oursnew.
%\input{tab/ablation_freq_text2}

% Preamble:
% \usepackage{pgfplots}
% \pgfplotsset{compat=1.18}

% Preamble:
% \usepackage{pgfplots}
% \pgfplotsset{compat=1.18}

\subsection{Discussion}
\label{sec:discussion}

\paragraph{Generalization across backbones and EO adaptation.}
Our benchmark suggests that both remote-sensing adaptation and general VLM quality matter for RSCIR. RS-adapted CLIP variants, such as RemoteCLIP, CLIP LAION-RS, and SkyCLIP-50, often improve over generic CLIP-style backbones under the same protocol. However, the gains from domain adaptation can be comparable to, or smaller than, the gains obtained by using stronger general-purpose backbones and training objectives, such as SigLIP. This indicates that advances in VLM architecture, pre-training data, and loss design transfer directly to composed retrieval in EO imagery. Consequently, future RSCIR systems should consider both EO-specific adaptation and improvements in general VLM representation quality.

\paragraph{Scalability and deployment trade-offs.}
The results show that training-free composition methods provide strong baselines for EO retrieval. FreeDom is the strongest method on PatternCom, while WeiCom and BASIC remain competitive lightweight alternatives. This is important for operational EO systems, where latency, scalability, and engineering complexity matter alongside accuracy. CIReVL can be effective, but requires image captioning and LLM-based rewriting, introducing additional computational cost and possible failure modes. In contrast, score-fusion or feature-calibration methods such as WeiCom and BASIC are simpler to deploy over large archives. Supervised multimodal heads such as MagicLens also transfer to EO imagery to some extent, but their performance may remain limited without training on remote-sensing compositional data.

\paragraph{Attribute-based and change-centric RSCIR require different mechanisms.}
The contrast between PatternCom and xView2-CIR shows that change-centric RSCIR should be treated as a distinct retrieval setting rather than a direct extension of attribute-based retrieval. PatternCom rewards methods that preserve class identity while injecting attribute-specific textual cues, which explains the strong performance of vocabulary-based methods such as FreeDom. In xView2-CIR, however, relevance requires the same scene/location and a target post-event state. This makes proxy-based visual expansion less reliable, because visually similar images often correspond to different locations and can dilute the scene-identity signal. These findings suggest that methods designed for attribute editing may not directly transfer to identity-preserving change retrieval.

\paragraph{Role in operational EO workflows.}
RSCIR is not a replacement for pixel-level change detection, damage assessment, or rapid mapping. Instead, it provides a complementary interface for controllable archive exploration and rapid evidence gathering. In practical workflows, composed retrieval can support location-aware search, retrieval of visually similar historical cases, analyst-in-the-loop exploration, quality control through hard-negative discovery, and weak supervision for downstream change models. This is especially useful when dense annotations, perfect co-registration, or complete temporal coverage are unavailable.

\paragraph{Interpretability and analyst interaction.}
Interpretability is also important for EO deployment. Methods based on explicit textual memories, such as FreeDom, provide human-readable surrogate concepts for the visual query, making it easier to inspect which semantic cues drive retrieval. Score-fusion methods such as WeiCom expose the relative contribution of image and text through the modality weight, while BASIC highlights the role of feature calibration and semantic projection. Such properties can help analysts diagnose whether failures arise from excessive visual dominance, weak text grounding, or scene-identity drift.

\paragraph{Sensitivity to text phrasing.}
Prompt phrasing is a practical factor in composed retrieval. On PatternCom, compact attribute modifiers such as \texttt{blue}, \texttt{water}, or \texttt{dense} are generally most effective, whereas verbose templates can dilute the attribute signal. In contrast, xView2-CIR can benefit from more descriptive, impact-oriented formulations, particularly for methods that rely more strongly on semantic text structure. This suggests that prompt design should be treated as part of the retrieval system: short modifiers are suitable for controlled attribute edits, while change-centric settings may require more explicit event or impact descriptions.
\section{Limitations}
\label{sec:limitations}

This study has several limitations. First, although xView2-CIR extends RSCIR toward operationally relevant change-centric retrieval, it remains relatively small and imbalanced, especially for rare disaster categories; results on these categories should therefore be interpreted as indicative rather than definitive. Second, our evaluation focuses on image-level retrieval and does not address finer-grained localization or dense temporal search. Third, some composition methods are more naturally tied to specific backbones or representation spaces, which limits strict cross-backbone comparability. Finally, prompt design and auxiliary vocabularies influence performance, particularly in change-centric scenarios, and require more systematic study. We therefore view this work as a benchmark-and-analysis foundation for RSCIR rather than a final solution to composed retrieval in Earth observation.
\section{Conclusion}
\label{sec:conclusions}

We studied remote sensing composed image retrieval (RSCIR), where a query combines a reference image with a textual modifier to express targeted retrieval intent. We established a unified benchmark on PatternCom with domain-grounded adaptations of representative composition methods and a standardized protocol spanning six vision--language backbones. To connect benchmarking with operational EO needs, we introduced xView2-CIR, a change-centric benchmark for disaster and damage monitoring, where retrieval requires both scene identity and a target post-event state.

Our experiments show that stronger vision--language backbones transfer directly to RSCIR, and that training-free composition strategies can be highly effective in EO settings. FreeDom achieves the strongest performance on PatternCom, while lightweight methods such as WeiCom and BASIC remain competitive and attractive for practical deployment. At the same time, xView2-CIR reveals that change-centric retrieval differs substantially from attribute-based editing: preserving scene identity changes the method ranking and exposes weaknesses in proxy-based expansion strategies.

Overall, our results position RSCIR as a practical complement to standard RSIR and change-analysis pipelines, especially when analysts need controllable, semantically guided access to large EO archives. Future work should expand change-centric benchmarks, study prompt and vocabulary design more systematically, and connect scene-level composed retrieval with finer-grained localization and temporal reasoning.

\section{Data and Code Availability Statement}
\label{sec:data}
The code and data supporting the findings of this study are publicly available at \url{https://github.com/billpsomas/rscir}. PatternCom is based on publicly available source data. The proposed xView2-CIR benchmark and the scripts used for data preparation, evaluation, and reproduction of the reported results are also provided in the repository.

\section{Acknowledgment}
\label{sec:acknowledgment}

This work was supported by the Czech Technical University in Prague grant No. SGS23/173/OHK3/3T/13, the EU Horizon Europe programme MSCA PF RAVIOLI (No. 101205297), and the Junior Star GACR GM 21-28830M. We acknowledge VSB–Technical University of Ostrava, IT4Innovati ons National Supercomputing Center, Czech Republic, for awarding this project (OPEN-33-67) access to the LUMI supercomputer, owned by the EuroHPC Joint Undertaking, hosted by CSC (Finland) and the LUMI consortium, through the Ministry of Education, Youth and Sports of the Czech Republic via the e-INFRA CZ project (ID: 90254). The access to the computational infrastructure of the OP VVV funded project CZ.02.1.01/0.0/0.0/16\_019/0000765 ``Research Center for Informatics'' is also gratefully acknowledged.

\bibliographystyle{cas-model2-names}
\bibliography{refs}

\clearpage
\appendix
\section*{Appendix}
\addcontentsline{toc}{section}{Appendix}
\section{Additional Experiments}
\label{app:exp}

\subsection{Dataset statistics}
\label{app:datasets}

\autoref{tab:dataset_patterncom} reports the detailed statistics of PatternCom, including the number of composed queries and positives for each attribute type, class, and target value. These statistics highlight the variability in the number of positives across attribute values, motivating the use of attribute-balanced macro-averaged mAP in the main evaluation.

\begin{castablehere}
\scriptsize
\renewcommand\theadfont{\scriptsize}
\centering
\setlength{\tabcolsep}{3pt}
\castablecaption{tab:dataset_patterncom}{\textbf{PatternCom statistics.} Breakdown by attribute type, class, and attribute value. \#Queries denotes the number of composed queries where the query text specifies the corresponding target value, and \#Positives denotes the number of database images that satisfy the relevance criterion (\emph{same class} and \emph{target attribute value}).}
\begin{tabular}{ccccc}
\toprule
\Th{Attribute} &
\Th{Class}  &
\Th{Value} &
\Th{\#Queries}   &
\Th{\#Positives} \\
\midrule

\multirow{11}{*}{color} & \multirow{2}{*}{airplane} & \texttt{white}  & 53  & 672 \\ 
&                         & \texttt{purple} & 672 & 53  \\ \cmidrule{2-5}

& \multirow{2}{*}{nursing home} & \texttt{white} & 383 & 85  \\ 
&                               & \texttt{gray}  & 85  & 383 \\ \cmidrule{2-5}

& \multirow{2}{*}{crosswalk} & \texttt{white}  & 388 & 412 \\
&                            & \texttt{yellow} & 412 & 388 \\ \cmidrule{2-5}

& \multirow{5}{*}{tennis court} & \texttt{blue}  & 287 & 339 \\
&                               & \texttt{brown} & 624 & 2   \\ 
&                               & \texttt{gray}  & 576 & 50  \\
&                               & \texttt{green} & 415 & 211 \\
&                               & \texttt{red}   & 602 & 24  \\ \midrule

\multirow{2}{*}{context} & \multirow{2}{*}{bridge} & \texttt{concrete} & 800 & 800 \\ 
&                         & \texttt{water}    & 800 & 800 \\ \midrule

\multirow{2}{*}{density} & \multirow{2}{*}{residential} & \texttt{sparse} & 800 & 800 \\
&                          & \texttt{dense}  & 800 & 800 \\ \midrule

\multirow{4}{*}{existence} & \multirow{2}{*}{parking} & \texttt{with cars}    & 653 & 947 \\ 
&                           & \texttt{without cars} & 947 & 653 \\ \cmidrule{2-5}
& \multirow{2}{*}{pier}     & \texttt{with boats}   & 255 & 1345 \\
&                           & \texttt{without boats}& 1345 & 255 \\ \midrule

\multirow{13}{*}{quantity} & \multirow{4}{*}{storage tank} & \texttt{one}   & 261 & 356 \\
&                           & \texttt{two}   & 498 & 119 \\ 
&                           & \texttt{three} & 552 & 65  \\
&                           & \texttt{four}  & 540 & 77  \\ \cmidrule{2-5}
& \multirow{4}{*}{wast. tr. plant} & \texttt{one}   & 78  & 724 \\
&                                  & \texttt{two}   & 758 & 44  \\ 
&                                  & \texttt{three} & 792 & 10  \\
&                                  & \texttt{four}  & 778 & 24  \\ \cmidrule{2-5}
& \multirow{5}{*}{basketball court} & \texttt{one}      & 383 & 340 \\
&                                   & \texttt{two}      & 437 & 286 \\ 
&                                   & \texttt{three}    & 702 & 21  \\
&                                   & \texttt{half}     & 662 & 61  \\
&                                   & \texttt{two-halfs}& 708 & 15  \\ \midrule

\multirow{7}{*}{shape} & \multirow{3}{*}{swimming pool} & \texttt{rectangular}   & 299 & 261 \\ 
&                        & \texttt{oval}          & 508 & 52  \\ 
&                        & \texttt{kidney-shaped} & 313 & 247 \\ \cmidrule{2-5}
& \multirow{2}{*}{river} & \texttt{curved}   & 623 & 177 \\
&                        & \texttt{straight} & 177 & 623 \\ \cmidrule{2-5}
& \multirow{2}{*}{road}  & \texttt{cross} & 800 & 800 \\
&                        & \texttt{round} & 800 & 800 \\

\bottomrule
\end{tabular}
%\label{tab:dataset_patterncom}
\end{castablehere}

\autoref{tab:dataset_xview} reports the statistics of xView2-CIR. This dataset should be viewed as a first evaluation benchmark for change-centric RSCIR rather than a complete operational disaster-monitoring dataset. Some disaster categories are small, which motivates reporting both macro-averaged and overall metrics and cautions against over-interpreting fine-grained differences on rare categories.

\begin{table}[htbp]
\scriptsize
\renewcommand\theadfont{\scriptsize}
\centering
\caption{\textbf{xView2-CIR statistics.}
Number of composed queries per disaster type and associated textual modifier. \#Queries denotes the number of pre-event composed queries. Each query has exactly one positive match (the post-event image of the \emph{same location} under the \emph{target disaster}), hence \#Positives $=1$ per query.}
\begin{tabular}{cccc}
\toprule
\Th{Disaster} &
\Th{Text query}  &
\Th{\#Queries}   &
\Th{\#Positives} \\
\midrule

hurricane  & \texttt{post-hurricane}  & 147 & 1 \\
wildfire   & \texttt{post-fire}       &  98 & 1 \\
flood      & \texttt{post-flood}      &  28 & 1 \\
tsunami    & \texttt{post-tsunami}    &   9 & 1 \\
earthquake & \texttt{post-earthquake} &   5 & 1 \\
volcano    & \texttt{post-volcano}    &   4 & 1 \\

\bottomrule
\end{tabular}

\label{tab:dataset_xview}
\end{table}

\subsection{Vision--language backbones}
\label{app:backbones}

We evaluate six vision--language models, all using a ViT-L/14 visual backbone:
\begin{itemize}

    \item \textbf{CLIP LAION-2B}: A CLIP model trained on LAION-2B~\cite{laion}, a dataset of 2.3 billion web-collected image--text pairs. We use the publicly released \texttt{laion2b\_s32b\_b82k} checkpoint from OpenCLIP~\cite{openclip}.

    \item \textbf{RemoteCLIP}~\cite{liu2023remoteclip}: A remote-sensing-adapted CLIP model initialized from OpenAI CLIP~\cite{clip} and fine-tuned on image--text pairs derived from annotated remote sensing datasets using synthetic captions.

    \item \textbf{OpenAI CLIP}~\cite{clip}: The original CLIP model released by OpenAI, trained on 400M web image--text pairs.

    \item \textbf{SigLIP}~\cite{siglip}: A vision--language model trained with a sigmoid-based contrastive loss instead of the standard softmax contrastive loss. We use the SigLIP model trained on WebLI.

    \item \textbf{CLIP LAION-RS}~\cite{skyscript2023}: A remote-sensing-adapted CLIP model initialized from OpenAI CLIP and fine-tuned on LAION-RS, a 726K-image remote sensing subset of LAION-2B.

    \item \textbf{SkyCLIP-50}~\cite{skyscript2023}: A remote-sensing-adapted CLIP model initialized from OpenAI CLIP and fine-tuned on SkyScript-50. SkyScript-50 contains 2.6M image--text pairs constructed by linking geo-referenced satellite imagery from Google Earth Engine~\cite{google_earth_engine} with OpenStreetMap~\cite{openstreetmap} annotations.

\end{itemize}

\subsection{Vocabulary construction}
\label{app:voc}

Some evaluated methods rely on a vocabulary or textual memory. To better align these methods with EO semantics, we generate a family of remote-sensing-specific vocabularies, denoted as RSText, using the prompt shown below. We use vocabularies of different sizes in the analysis and construct HybridText-23k by merging RSText-2k with Open Images v7.

\begin{tcolorbox}[colback=blue!5!white,colframe=blue!75!black,title=Vocabulary generation prompt]
\label{box:corpus_prompt}
I need a fine-grained, diverse vocabulary list suitable for detailed land-use/land-cover (LULC), remote sensing, and object-detection datasets. The vocabulary should be explicitly descriptive, detailed, and balanced across multiple thematic categories, including but not limited to:
\begin{itemize}
    \item Urban infrastructure
    \item Cultural and historical sites
    \item Recreational and tourism areas
    \item Transportation and logistics
    \item Construction and housing
    \item Education and healthcare
    \item Natural features and ecosystems
    \item Biodiversity and wildlife habitats
    \item Sustainability initiatives and eco-friendly technologies
    \item Agriculture, farming, and traditional land-use practices
    \item Marine and aquatic environments
\end{itemize}

Each vocabulary entry should ideally be short (1--4 words), clear, explicit, self-contained, and suitable for remote sensing imagery annotation. Provide vocabulary entries systematically in batches of 100, carefully ensuring thematic diversity and granularity until a total of at least 2000 entries is reached.
\end{tcolorbox}

\subsection{Additional details for adapted methods}
\label{app:implementation}

This section provides additional implementation details for the adapted composed image retrieval methods evaluated in the main paper. Unless otherwise stated, we follow the official implementations or default settings from the original papers and introduce only the minimal adaptations needed for EO imagery.

\paragraph{CompoDiff} is a compositional diffusion framework~\cite{stablediffusion} that samples an image-conditioned text embedding via a denoising diffusion process in the joint CLIP embedding space. Sampling is conditioned on (i) the query image embedding, (ii) the composed text prompt (\eg, \texttt{red \{*\}}), and (iii) a negative prompt (\eg, \texttt{low quality}). We evaluate CompoDiff in its native configuration and, where applicable, under the hybrid protocol described above. To better preserve identity/class information, we blend the sampled embedding with the original image embedding using a source weight parameter. We report results with 10 diffusion steps and source weight $0.4$.

\paragraph{Pic2Word} learns a lightweight MLP that maps image embeddings to a textual embedding suitable for composition, enabling image-to-text inversion with a single forward pass. For retrieval, we use the template \texttt{\{attribute\_value\} \{*\}}, where \texttt{*} denotes the inverted image token. We evaluate Pic2Word in its standard setting and, when relevant, under the hybrid protocol described above.

\paragraph{SEARLE} performs test-time textual inversion by optimizing a pseudo token such that its text-encoder embedding matches the query image embedding, regularized by an LLM-guided loss. To ground SEARLE in RS, we replace the Open Images v7~\cite{open_images} vocabulary with our domain-specific HybridText-23k vocabulary and use the RS-adapted prompt template \texttt{a satellite image of \{concept\} that}. For retrieval, we use the composed text template \texttt{\{attribute\_value\} \{\$\}}, where the attribute value (\eg, \texttt{red}) modifies the learned placeholder token \texttt{\$}. We report results with 200 optimization steps, learning rate $0.002$, and GPT-loss weight $\lambda_{\text{GPT}}{=}0.25$.

\paragraph{CIReVL} is a caption-guided composition pipeline that converts the query image into text and then edits it to reflect the modifier. It consists of: (i) captioning the query image, (ii) editing the caption with an instruction-tuned LLM given the textual modifier, and (iii) encoding the edited description with the retrieval backbone for text-to-image search. We use BLIP-2~\cite{blip2} (blip2-opt-2.7b) as the captioner and LLaMA-2 7B~\cite{llama2} (Llama-2-7b-chat-hf) as the editor. To ground CIReVL to PatternCom, we design a domain-specific modifier prompt covering the six attribute types to enforce faithful attribute changes while preserving scene semantics.

%-----------------------------------------------
\begin{tcolorbox}[colback=blue!5!white,colframe=blue!75!black,title=CIReVL modifier prompt for PatternCom]
You are given a high-resolution satellite image caption describing the ``Image Content'', along with a single-word ``Instruction'' that specifies a modification to apply to the scene.
Generate a complete, natural-language ``Edited Description'' that integrates the modification while preserving all other aspects of the original content.
The instruction will always belong to one of six attribute types: color, context, density, existence, quantity, shape. Example (shape):
\begin{itemize}
\item Image Content: a satellite image of a rectangular swimming pool in a resort.
\item Instruction: \texttt{oval}
\item Edited Description: a satellite image of an \emph{oval} swimming pool in a resort.
\end{itemize}
\end{tcolorbox}
%-----------------------------------------------
\paragraph{MagicLens} augments an OpenAI CLIP backbone with a transformer based compositional head trained for image--text composition under supervised triplets. We use the official pretrained model (ViT-L/14) and apply it directly to both queries and database images, without additional adaptation. Retrieval is performed using the joint embeddings produced by the MagicLens head.
%-----------------------------------------------
\paragraph{WeiCom} is a training-free score-fusion approach that combines image-to-image and text-to-image similarities after calibrating them to a comparable scale. We report results with $\lambda{=}0.5$, corresponding to equal modality contribution.
%-----------------------------------------------
\paragraph{BASIC} is a training-free composition method that combines image-to-image and text-to-image similarities after centering and semantic projection, optionally applying image-side query expansion, and fusing modalities with a Harris-regularized multiplicative score. To ground BASIC in RS, we augment $C_{+}$ with RSText-150. We report results using $250$ principal components for the semantic projection, contrastive scaling $\alpha{=}0.2$, query expansion with $25$ nearest neighbors, and Harris regularization weight $\lambda_{\text{Harris}}{=}0.1$.
%-----------------------------------------------
\paragraph{FreeDom} is a memory-based textual inversion method that constructs an interpretable text representation of the query image using a large vocabulary and proxy-image expansion. Given a query image and text, it first retrieves $k$ proxy images via image similarity, then retrieves the top-$n$ vocabulary labels per proxy, and retains the most frequent $m$ labels overall. Each retained label $w$ is composed with the modifier using \texttt{\{attribute\_value\} \{w\}}, and the resulting text embeddings are fused by frequency-weighted averaging to form the final query representation. We ground the textual memory to remote sensing using HybridText-23k. We report results using $k{=}20$, $n{=}7$, and $m{=}7$.

The evaluated methods span a broad deployment spectrum: some are fully training-free and lightweight at inference (\eg, WeiCom, BASIC), some require large auxiliary vocabularies or proxy retrieval (\eg, FreeDom), some rely on captioning and LLM editing (\eg, CIReVL), and others incur heavier per-query optimization or sampling cost (\eg, SEARLE, CompoDiff). This diversity is important when assessing their suitability for operational EO archives.
%-----------------------------------------------

\begin{casfigurehere}
\centering
\begin{adjustbox}{max width=\linewidth}
\begin{tikzpicture}
\begin{axis}[
    width=9cm,
    height=4.5cm,
    ylabel={mAP (\%)},
    xlabel={Hyperparameters ($k{=}1$, varying $m{=}n$)},
    xtick=data,
    symbolic x coords={k1_mn1, k1_mn3, k1_mn5, k1_mn7, k1_mn10, k1_mn15},
    xticklabels={
        $1$,
        $3$,
        $5$,
        $7$,
        $10$,
        $15$
    },
    xticklabel style={anchor=north},
    ymin=0, ymax=40,
    xmajorgrids=true,
    ymajorgrids=true,
    grid style={gray!10},
    legend style={
      at={(1.02,0.99)},
      anchor=north west,
      draw=black,
      fill=white,
      fill opacity=0.9,
      text opacity=1,
      font=\scriptsize
    },
    legend cell align={left},
    %every axis plot/.append style={line width=1.2pt, solid}
    %every node near coord/.append style={font=\tiny},
    %nodes near coords,
    tick label style={font=\scriptsize},
    label style={font=\scriptsize},
]

% Attribute-wise plots
\addplot+[mark=*, thick, line width=1.2pt, color=blue!55, mark options={fill=blue!55}] coordinates {
    (k1_mn1, 9.93) (k1_mn3, 14.66) (k1_mn5, 16.55)
    (k1_mn7, 18.01) (k1_mn10, 18.26) (k1_mn15, 17.95)
};
\addlegendentry{Hurricane}

\addplot+[mark=*, thick, line width=1.2pt,color=orange!65, mark options={fill=orange!65}] coordinates {
    (k1_mn1, 8.46) (k1_mn3, 14.30) (k1_mn5, 16.36)
    (k1_mn7, 17.46) (k1_mn10, 16.92) (k1_mn15, 14.93)
};
\addlegendentry{Wildfire}

\addplot+[mark=*, thick, line width=1.2pt,color=ForestGreen!55, mark options={fill=ForestGreen!55}] coordinates {
    (k1_mn1, 5.41) (k1_mn3, 4.6) (k1_mn5, 4.23)
    (k1_mn7, 3.77) (k1_mn10, 3.59) (k1_mn15, 4.45)
};
\addlegendentry{Flood}

\addplot+[mark=*, thick, line width=1.2pt,color=red!55, mark options={fill=red!55}] coordinates {
    (k1_mn1, 34.01) (k1_mn3, 33.34) (k1_mn5, 35.04)
    (k1_mn7, 35.03) (k1_mn10, 36.68) (k1_mn15, 32.05)
};
\addlegendentry{Tsunami}

\addplot+[mark=*, thick, line width=1.2pt,color=purple!55, mark options={fill=purple!55}] coordinates {
    (k1_mn1, 0.95) (k1_mn3, 2.04) (k1_mn5, 3.22)
    (k1_mn7, 3.21) (k1_mn10, 3.4) (k1_mn15, 3.81)
};
\addlegendentry{Earthquake}

\addplot+[mark=*, thick, solid, line width=1.2pt,color=teal!55, mark options={fill=teal!55}] coordinates {
    (k1_mn1, 10.08) (k1_mn3, 15.12) (k1_mn5, 19.94)
    (k1_mn7, 25.88) (k1_mn10, 26.25) (k1_mn15, 23.8)
};
\addlegendentry{Volcano}

\addplot+[mark=*, thick, solid, line width=1.2pt,color=black!55, mark options={fill=black!55}] coordinates {
    (k1_mn1, 9.59) (k1_mn3, 13.94) (k1_mn5, 15.69)
    (k1_mn7, 16.84) (k1_mn10, 16.82) (k1_mn15, 15.91)
};
\addlegendentry{Total}

\end{axis}
\end{tikzpicture}
\end{adjustbox}
\vspace{-16pt}
\casfigcaption{fig:freedom_freq_expansion_xview2}{\textbf{Impact of frequency-based textual expansion} on FreeDom with OpenAI CLIP on xView2-CIR with $k{=}1$ and increasing $m{=}n$. Aggregating multiple textual labels improves mAP across most disaster types, especially hurricane and wildfire. Performance peaks at $m{=}n{=}7$, after which further expansion yields diminishing returns.}
%\label{fig:freedom_freq_expansion_xview2}
\end{casfigurehere}

\subsection{Additional sensitivity and ablation analyses}
\label{app:additional_ablations}

\paragraph{Textual expansion on xView2-CIR.}
\autoref{fig:freedom_freq_expansion_xview2} complements the main-paper analysis of FreeDom textual expansion by reporting results on xView2-CIR. Visual expansion is disabled ($k{=}1$), while the number of aggregated textual concepts is varied as $m{=}n$. Moderate textual expansion improves performance, with the best Total score obtained at $m{=}n{=}7$. Larger expansions provide diminishing returns, suggesting that overly long textual surrogates can introduce noisy or off-target concepts.

%-----------------------------------------------
\paragraph{CompoDiff hyperparameter sensitivity.}
\autoref{fig:compodiff_hparam_sweep} analyzes CompoDiff sensitivity to the number of diffusion timesteps $t$ and the source weight used to mix the reference image with the generated proxy. Performance improves when moving from very small source weights to the mid-range ($0.3$--$0.5$), indicating that both the original image and the generative signal are useful for composition. Moderate diffusion budgets ($t{=}10$--$20$) are generally sufficient, while larger budgets provide limited additional gains.

\begin{casfigurehere}
\centering
\begin{tikzpicture}
\begin{axis}[
    width=0.70\linewidth,
    height=4cm,
    xlabel={Source weight},
    ylabel={Average mAP (\%)},
    xmin=0.1, xmax=0.6,
    xtick={0.1,0.2,0.3,0.4,0.5,0.6},
    ymajorgrids=true,
    xmajorgrids=true,
    grid style={gray!10},
    legend style={at={(0.65,0.52)}, anchor=north west, draw=black, fill=white, font=\tiny},
    legend cell align=left,
    tick label style={font=\scriptsize},
    label style={font=\scriptsize},
]

% timesteps = 10
\addplot+[mark=*, mark size=1.2pt, line width=1.2pt, color=blue!55, mark options={fill=blue!55}] coordinates {
    (0.1,  8.613333333)
    (0.2, 13.48333333)
    (0.3, 16.80333333)
    (0.4, 17.55833333)
    (0.5, 16.955)
    (0.6, 16.19666667)
};
\addlegendentry{$t=10$}

% timesteps = 20
\addplot+[mark=*, mark size=1.2pt, line width=1.2pt, color=orange!55, mark options={fill=orange!55}] coordinates {
    (0.1,  7.313333333)
    (0.2, 12.06833333)
    (0.3, 15.55833333)
    (0.4, 16.85)
    (0.5, 16.555)
    (0.6, 16.00333333)
};
\addlegendentry{$t=20$}

% timesteps = 30
\addplot+[mark=*, mark size=1.2pt, line width=1.2pt, color=red!55, mark options={fill=red!55}] coordinates {
    (0.1,  7.213333333)
    (0.2, 11.835)
    (0.3, 15.38)
    (0.4, 16.565)
    (0.5, 16.45333333)
    (0.6, 15.97166667)
};
\addlegendentry{$t=30$}

\end{axis}
\end{tikzpicture}
\vspace{-6pt}
\casfigcaption{fig:compodiff_hparam_sweep}{\textbf{CompoDiff hyperparameter sweep} with SkyCLIP-50 on PatternCom. We report average mAP as a function of source weight for different numbers of diffusion timesteps.}
%\label{fig:compodiff_hparam_sweep}
\end{casfigurehere}
%-----------------------------------------------
\paragraph{SEARLE hyperparameter sensitivity.}
\autoref{fig:searle_hparam_sweep} evaluates SEARLE under different learning rates, inversion steps, and GPT mixing coefficients $\lambda_{\mathrm{GPT}}$. Performance is relatively stable around $\lambda_{\mathrm{GPT}}{=}0.25$ and moderate learning rates, while very small learning rates degrade results. Larger GPT mixing does not provide consistent gains. Overall, SEARLE is moderately sensitive to optimization settings but remains stable around prior-work defaults.

\begin{casfigurehere}
\centering
\begin{tikzpicture}
\begin{axis}[
    width=0.65\linewidth,
    height=4cm,
    xlabel={Learning rate},
    ylabel={Average mAP (\%)},
    xmode=log,
    xmin=0.00015, xmax=0.03,
    xtick={0.0002,0.002,0.007,0.02},
    xticklabels={0.0002, 0.002, 0.007, 0.02},
    ymajorgrids=true,
    xmajorgrids=true,
    grid style={gray!10},
    legend style={at={(0.40,0.63)}, anchor=north west, draw=black, fill=white, font=\tiny},
    legend cell align=left,
    tick label style={font=\scriptsize},
    label style={font=\scriptsize},
]

% lambda_GPT = 0.25 (steps=200)
\addplot+[mark=*, mark size=1.2pt, line width=1.2pt, color=blue!55, mark options={fill=blue!55}] coordinates {
    (0.02,   12.47)
    (0.007,  12.57)
    (0.002,  12.57666667)
    (0.0002, 11.89166667)
};
\addlegendentry{$\lambda_{\text{GPT}}=0.25$ (steps=200)}

% lambda=0.5, steps=200  (single point -> only mark)
\addplot+[only marks, mark=*, mark size=1.2pt, color=orange!65, mark options={fill=orange!65}]
coordinates {(0.007, 11.89166667)};
\addlegendentry{$\lambda_{\text{GPT}}=0.5$ (steps=200)}

% lambda=0.5, steps=350  (single point -> only mark)
\addplot+[only marks, mark=*, mark size=1.2pt, color=red!65, mark options={fill=red!65}]
coordinates {(0.02, 12.37)};
\addlegendentry{$\lambda_{\text{GPT}}=0.5$ (steps=350)}

\end{axis}
\end{tikzpicture}
\vspace{-6pt}
\casfigcaption{fig:searle_hparam_sweep}{\textbf{SEARLE hyperparameter sweep} with CLIP LAION-2B on PatternCom. We report average mAP as a function of learning rate, number of inversion steps, and GPT mixing coefficient $\lambda_{\mathrm{GPT}}$.}
%\label{fig:searle_hparam_sweep}
\end{casfigurehere}
%-----------------------------------------------
\paragraph{Ablation of BASIC components}
\autoref{tab:basic_ablation_components} and~\ref{tab:basic_ablation_components_xview2} analyze the contribution of BASIC components on PatternCom and xView2-CIR. Across both datasets, removing \emph{centering} or \emph{semantic projection} causes the largest drops, showing that modality calibration and semantic subspace alignment are the main drivers of BASIC's performance. Other components, including Harris weighting, min-based normalization, text contextualization, and query expansion, provide smaller and more dataset-dependent gains. Notably, query expansion is harmful on xView2-CIR, where relevance requires the same scene/location and visually similar proxies often introduce off-location evidence. Overall, BASIC's gains primarily come from representation calibration and semantic projection, while auxiliary heuristics play a secondary role.

\begin{table*}[t]
\centering
\scriptsize
\setlength{\tabcolsep}{4pt}
\caption{\textbf{BASIC component ablation on PatternCom with OpenAI CLIP.} We report attribute-wise and average mAP (\%) when removing components from the full pipeline: mean centering (\textbf{Cen.}), min-based normalization (\textbf{Norm.}), Harris weighting (\textbf{Har.}), text contextualization (\textbf{Con.}), semantic projection (\textbf{Proj.}), and query expansion (\textbf{Q. Ex.}).}
%\vspace{-4pt}
\label{tab:basic_ablation_components}
\begin{tabular}{cccccc|cccccc|c}
\toprule
\textbf{Cen.} & \textbf{Norm.} & \textbf{Har.} & \textbf{Con.} & \textbf{Proj.} & \textbf{Q. Ex.} &
Color & 
Context & 
Density & 
Existence & 
Quantity & 
Shape & 
Avg. \\
\midrule
\cmark & \cmark & \cmark & \cmark & \cmark & \cmark &
38.31 & 24.53 & 26.35 & 30.94 & 18.03 & 35.73 & \best{28.98} \\
\midrule

\cmark & \cmark & \xmark & \cmark & \cmark & \cmark &
36.69 & 22.87 & 24.34 & 28.40 & 16.14 & 32.74 & 26.86 \\

\cmark & \xmark & \xmark & \cmark & \cmark & \cmark &
26.57 & 30.51 & 24.63 & 11.93 & 26.08 & 23.27 & 23.83 \\

\cmark & \cmark & \cmark & \xmark & \cmark & \cmark &
34.60 & 24.56 & 19.17 & 12.81 & 26.44 & 17.03 & 22.44 \\

\cmark & \cmark & \cmark & \cmark & \xmark & \cmark &
23.22 & 24.41 & 18.97 & 3.11 & 24.40 & 20.19 & 19.05 \\

\xmark & \cmark & \cmark & \cmark & \cmark & \cmark &
28.90 & 18.95 & 22.77 & 8.02 & 12.57 & 16.33 & 17.92 \\

\cmark & \cmark & \cmark & \cmark & \cmark & \xmark &
37.16 & 24.51 & 24.93 & 28.62 & 16.57 & 34.22 & 27.67 \\
\bottomrule
\end{tabular}

\end{table*}

\begin{table*}[t]
\centering
\scriptsize
\setlength{\tabcolsep}{4pt}
\caption{\textbf{BASIC component ablation on xView2-CIR with OpenAI CLIP.} We report disaster-wise mAP (\%), macro-average mAP (Avg.), and overall mAP (Total) when removing components from the full pipeline. Component abbreviations follow \autoref{tab:basic_ablation_components}.}
%\vspace{-4pt}
\label{tab:basic_ablation_components_xview2}
\begin{tabular}{cccccc|cccccc|c|c}
\toprule
\textbf{Cen.} & \textbf{Norm.} & \textbf{Har.} & \textbf{Con.} & \textbf{Proj.} & \textbf{Q. Ex.} &
Hurricane & 
Fire & 
Flood & 
Tsunami & 
Earthquake & 
Volcano & 
Avg. & 
Total \\
\midrule
\cmark & \cmark & \cmark & \cmark & \cmark & \cmark &
9.36 & 7.74 & 7.54 & 19.52 & 1.18 & 13.96 & 9.88 & 8.88 \\

\cmark & \cmark & \cmark & \cmark & \cmark & \xmark &
22.71 & 18.49 & 14.59 & 46.15 & 1.69 & 59.20 & \best{27.14} & \best{21.38} \\
\midrule

\cmark & \cmark & \xmark & \cmark & \cmark & \xmark &
21.16 & 17.45 & 14.60 & 47.63 & 1.79 & 51.04 & 25.61 & 20.18 \\

\cmark & \xmark & \xmark & \cmark & \cmark & \xmark &
16.39 & 17.08 & 11.11 & 29.13 & 1.14 & 64.17 & 23.17 & 16.90 \\

\cmark & \cmark & \cmark & \xmark & \cmark & \xmark &
18.48 & 16.69 & 14.38 & 32.37 & 2.62 & 57.25 & 23.63 & 18.17 \\

\cmark & \cmark & \cmark & \cmark & \xmark & \xmark &
16.03 & 14.74 & 13.68 & 24.25 & 1.26 & 59.01 & 21.49 & 15.96 \\

\xmark & \cmark & \cmark & \cmark & \cmark & \xmark &
0.68 & 0.71 & 1.24 & 17.02 & 1.39 & 6.29 & 4.56 & 1.34 \\

\bottomrule
\end{tabular}

\end{table*}
%-----------------------------------------------
\subsection{Sensitivity to text query phrasing}
\label{app:prompt_sensitivity}

We study how composed retrieval methods respond to alternative textual formulations of the modifier. For PatternCom, we test three templates applied to each attribute value $v$: R1 uses a \emph{being} formulation (\eg, \texttt{being blue}, \texttt{being rectangular-shaped}); R2 uses a \emph{having} formulation (\eg, \texttt{having blue color}, \texttt{having rectangular shape}); and R3 uses longer, relational descriptions such as \texttt{with the main object modified to have blue color} or \texttt{with the main object modified to have rectangular sha pe}. As shown in~\autoref{tab:patterncom_rephrasing}, the original plain attribute word (\eg, \texttt{blue}, \texttt{water}, \texttt{dense}) is generally the most reliable choice, especially for stronger multimodal methods. More verbose templates tend to reduce performance, suggesting that additional function words can dilute the compact attribute signal.

For xView2-CIR, the plain query is \texttt{post-\{disaster\}}, while R1 uses impact-oriented rephrasing (\eg, \texttt{burned area}, \texttt{flooded area}) and R2 uses disaster-explicit formulations (\eg, \texttt{fire-affected region}, \texttt{seismic damag e}). As shown in~\autoref{tab:xview_rephrasing}, prompt sensitivity is more method-dependent in this change-centric setting. More descriptive formulations can benefit methods that rely more strongly on semantic text structure, particularly BASIC, while other methods are more sensitive to distribution shifts introduced by rewording. Overall, these results indicate that prompt design is dataset- and method-dependent: compact modifiers are effective for controlled attribute edits, whereas change-centric retrieval may benefit from more explicit event or impact descriptions.

\begin{table}[ht]
\centering
\scriptsize
\begin{minipage}[t]{0.48\textwidth}
\centering
\setlength{\tabcolsep}{4pt}
\captionof{table}{\textbf{Text query rephrasing on PatternCom} with CLIP LAION-2B. We report average mAP (\%) for the original modifier (Plain) and three rephrased variants (R1--R3).}
%\vspace{-6pt}
\label{tab:patterncom_rephrasing}

\begin{tabular}{l@{\hspace{4pt}}|@{\hspace{4pt}}c@{\hspace{4pt}}@{\hspace{4pt}}c@{\hspace{4pt}}@{\hspace{4pt}}c@{\hspace{4pt}}@{\hspace{4pt}}c@{\hspace{4pt}}}
\toprule
\multirow{2}{*}{\Th{Method}} &
\multicolumn{4}{c@{\hspace{4pt}}@{\hspace{4pt}}}{Average} \\
\cmidrule(l{0pt}r{8pt}){2-5}
 & Plain & R1 & R2
 & R3 \\
\midrule
Text-only               & 5.57 & \best{9.16} & 7.07 & 3.52  \\
Text$+$Image          & 16.41 & \best{17.70} & 17.48 & 15.10  \\
Text$\times$Image     & 21.00 & 19.03 & \best{21.58} & 16.76  \\
\midrule
WeiCom                  & 21.75 & 20.86 & \best{23.31} & 16.38  \\
BASIC                   & 16.47 & \best{17.50} & 17.73 & 17.34 \\
FreeDom                 & \best{38.59} & 33.06 & 31.48 & 29.53 \\
\bottomrule
\end{tabular}

\end{minipage}
\hfill
\begin{minipage}[t]{0.48\textwidth}
\centering
\setlength{\tabcolsep}{0.5pt}
\captionof{table}{\textbf{Text query rephrasing on xView2-CIR} with OpenAI CLIP. We report macro-average mAP (Average) and overall mAP (Total) for the original query (Plain) and two rephrased variants (R1--R2).}
%\vspace{-6pt}
\label{tab:xview_rephrasing}
\begin{tabular}{l@{\hspace{4pt}}|@{\hspace{4pt}}c@{\hspace{4pt}}@{\hspace{4pt}}c@{\hspace{4pt}}@{\hspace{4pt}}c@{\hspace{4pt}}|@{\hspace{4pt}}c@{\hspace{4pt}}@{\hspace{4pt}}c@{\hspace{4pt}}@{\hspace{4pt}}c}
\toprule
\multirow{2}{*}{\Th{Method}} &
\multicolumn{3}{c@{\hspace{4pt}}|@{\hspace{4pt}}}{Average} &
\multicolumn{3}{c}{Total} \\
\cmidrule(l{0pt}r{8pt}){2-4}
\cmidrule(l{0pt}r{4pt}){5-7}
 & Plain & R1 & R2
 & Plain & R1 & R2 \\
\midrule
Text-only               & \best{7.57} & 4.53 & 4.30 & \best{2.97} & 1.76 & 2.06 \\
%Image-only              & 9.00 & 9.00 & 9.00 & 5.53 & 5.53 & 5.53 \\
Text$+$Image          & \best{17.73} & 15.27 & 15.63 & \best{12.14} & 9.14 & 10.22 \\
Text$\times$Image     & \best{25.94} & 14.97 & 12.13 & \best{16.35} & 10.38 & 9.65 \\
\midrule
WeiCom                  & \best{30.67} & 24.39 & 19.22 & \best{21.40} & 12.58 & 13.13 \\
BASIC                   & \best{27.14} & 17.39 & 17.36 & \best{21.38} & 18.56 & 18.17 \\
FreeDom                 & \best{17.23} & 9.86 & 11.98 & \best{16.84} & 10.83 & 12.19 \\
\bottomrule
\end{tabular}

\end{minipage}
\end{table}

\end{document}